%% file: main.tex
\documentclass[10pt,twocolumn,letterpaper]{article}

\usepackage{cvpr}              

\usepackage{graphicx}
\usepackage{amsmath}
\usepackage{amssymb}
\usepackage{booktabs}
\usepackage{tabularx}
\usepackage{float}
\usepackage[resetlabels]{multibib}
\usepackage[accsupp]{axessibility}  

\usepackage[pagebackref,breaklinks,colorlinks]{hyperref}

\usepackage[capitalize]{cleveref}
\crefname{section}{Sec.}{Secs.}
\Crefname{section}{Section}{Sections}
\Crefname{table}{Table}{Tables}
\crefname{table}{Tab.}{Tabs.}

\newcites{supp}{References [Supplementary]}%
\newcommand\Mark[1]{\textsuperscript#1}

\def\cvprPaperID{4} 
\def\confName{CVPR}
\def\confYear{2022}

\input{preamble}

\begin{document}
\input{sec/0_metadata}
\maketitle
\input{sec/0_abstract}
\input{sec/1_introduction}

\input{sec/2_related}

\input{sec/3_motivation}
\input{sec/4_1_h1}
\input{sec/4_3_h3}

\input{sec/4_2_h2}

\input{sec/5_discussion}
\input{sec/6_conclusions}

{
    \small
    \bibliographystyle{ieee_fullname}
    \bibliography{macros,main}
}

\input{sec/X_supplementary}

\clearpage
\pagebreak
{
    \small
    \bibliographystylesupp{alpha}
    \bibliographysupp{main_supp}
}


\end{document}

%% file: preamble.tex
\usepackage{overpic}
\usepackage{enumitem} 
\usepackage{overpic} 
\usepackage{color}

\definecolor{turquoise}{cmyk}{0.65,0,0.1,0.3}
\definecolor{purple}{rgb}{0.65,0,0.65}
\definecolor{dark_green}{rgb}{0, 0.5, 0}
\definecolor{orange}{rgb}{0.8, 0.6, 0.2}
\definecolor{red}{rgb}{0.8, 0.2, 0.2}
\definecolor{darkred}{rgb}{0.6, 0.1, 0.05}
\definecolor{blueish}{rgb}{0.0, 0.3, .6}
\definecolor{light_gray}{rgb}{0.7, 0.7, .7}
\definecolor{pink}{rgb}{1, 0, 1}
\definecolor{greyblue}{rgb}{0.25, 0.25, 1}

\usepackage{blindtext}

\renewcommand{\paragraph}[1]{\vspace{0.5em}\noindent\textbf{#1}:\hspace{0.5em}}

%% file: sec/0_metadata.tex
\title{What’s in a Caption? Dataset-Specific Linguistic Diversity and Its Effect on Visual Description Models and Metrics}

\author{David M. Chan\Mark{1}, Austin Myers\Mark{2}, Sudheendra Vijayanarasimhan\Mark{2}, David A. Ross\Mark{2}\\
Bryan Seybold\Mark{2}, John F. Canny\Mark{1}\\
\Mark{1}University of California, Berkeley $\quad$ \Mark{2}Google Research\\
{\tt\small \{davidchan, canny\}@berkeley.edu}\\
{\tt\small \{aom, svnaras, dross, seybold\}@google.com}
}

%% file: sec/0_abstract.tex
\begin{abstract}
While there have been significant gains in the field of automated video description, the generalization performance of automated description models to novel domains remains a major barrier to using these systems in the real world. Most visual description methods are known to capture and exploit patterns in the training data leading to evaluation metric increases, but what are those patterns? In this work, we examine several popular visual description datasets, and capture, analyze, and understand the dataset-specific linguistic patterns that models exploit but do not generalize to new domains. At the token level, sample level, and dataset level, we find that caption diversity is a major driving factor behind the generation of generic and uninformative captions. We further show that state-of-the-art models even outperform held-out ground truth captions on modern metrics, and that this effect is an artifact of linguistic diversity in datasets. Understanding this linguistic diversity is key to building strong captioning models, we recommend several methods and approaches for maintaining diversity in the collection of new data, and dealing with the consequences of limited diversity when using current models and metrics.
\end{abstract}

%% file: sec/1_introduction.tex
\section{Introduction}
\label{sec:intro}

\begin{figure}
    \centering
    \includegraphics[width=\linewidth]{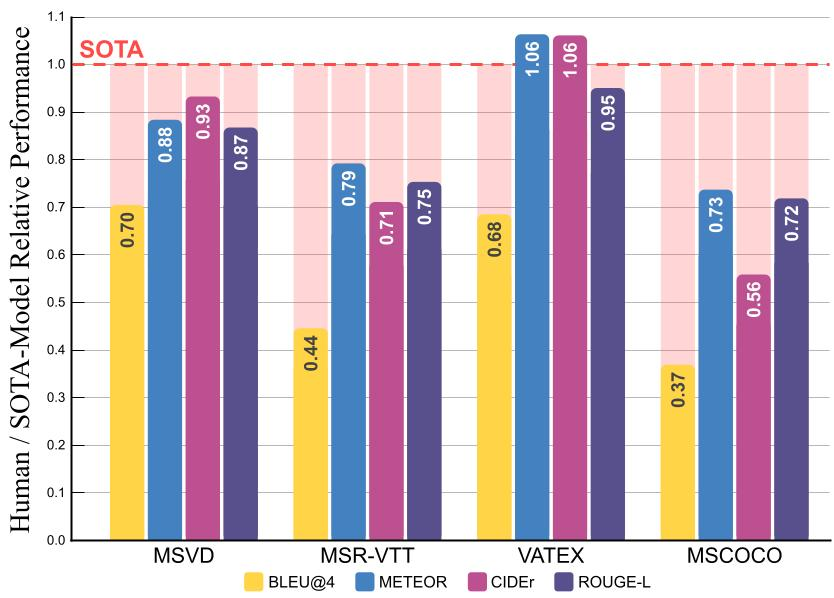}
    \caption{Captions generated by state-of-the-art (SOTA) models outperform held-out ground truth captions written by humans on common visual description datasets and metrics.
    Despite being far from human-level, SOTA models appear to outperform humans on most datasets and metrics, with the exception of VATEX, a relatively new dataset (and not even on all metrics).
    This discrepancy begs the question, ``What causes these effects?" and ``Are these effects indicative of a more serious issue with visual description datasets or model evaluation methods?" The figure above shows metric performance normalized to a recent SOTA model across several visual description datasets.
    }
    \label{fig:motivation}
\end{figure}

Automated visual description is an emergent field in computer vision, aiming to generate natural language descriptions of visual information. With various applications including digital accessibility \cite{youdescribe} and video summarization \cite{zhang16} as well as indexing and search \cite{miech2019howto100m}, methods for visual description have the potential to impact the daily lives of billions of users. Recent improvements such as vision and language pre-training \cite{zhang2021vinvl}, compositional and graph methods \cite{perez2021improving, chen2021motion}, and non-autoregressive training \cite{liu2021o2na} have driven metric performance on standard benchmarks such as MSR-VTT \cite{xu2016msr} and MS-COCO \cite{lin2014microsoft} to new heights.

Unfortunately, despite recent improvements in model architectures \cite{liu2021o2na, perez2021improving}, metrics \cite{jiang2019tiger, wang2021faier}, and datasets \cite{monfort2021spoken, wang2019vatex}, automated visual description has been plagued by issues of poor generalization and description quality \cite{stefanini2021show, aafaq2019video, smeaton2019exploring, yang2020towards}. Models consistently perform poorly on novel data, generate nonsense descriptions, or produce descriptions that are too vague to be of use to visually impaired users \cite{macleod2017understanding}. It remains an open question in visual description to understand the source of these generalization issues.

This paper is motivated by both the fact that often state-of-the-art methods outperform leave-one-out experiments with ground truth sample data (explored in \ref{sec:motivation}) as well as results demonstrating poor cross-dataset generalization in video captioning from Smeaton \etal \cite{smeaton2019exploring} and Yang \etal \cite{yang2021deconfounded}. We find that one major issue in current datasets---description linguistic diversity---explains a great deal about model evaluations.

Our work, consisting of analyses on several popular visual description datasets, contains several primary contributions:
\begin{enumerate}
    \item We demonstrate that a lack of linguistic diversity at a token and n-gram level can bias models to generate descriptions lacking in semantic detail (\autoref{sec:h1}).
    \item We show that diversity among ground truths for a single visual context presents a catch-22: low within-sample linguistic diversity leads to generic captions, as information is repetitive; on the other hand, high within-sample diversity leads to a breakdown of single-sample metrics, causing inconsistencies in model evaluation and inaccurate understanding of model performance (\autoref{sec:h3}).
    \item We detail how a lack of semantic diversity at the dataset level can encourage models to generate generic descriptions through classification, instead of learning to understand and relay visual phenomena at various levels of detail (\autoref{sec:h2}).
    \item We discuss our findings demonstrating the need for future research in visual description datasets, methods, and metrics, present recommendations on possible solutions to current linguistic diversity, and introduce a new toolkit for dataset evaluation and split generation focused on linguistic diversity (\autoref{sec:discussion}).
\end{enumerate}

%% file: sec/2_related.tex
\section{Experimental Design}

In this work, we explore the field of visual description data through the lens of some of the most popular visual description datasets. While there are a large number of visual description datasets to choose from, we decided to focus on some of the most common datasets for video description, and an additional dataset for image description: \footnote{As described in \autoref{sec:discussion}, we make the tools available for this analysis public, so any additional datasets can be analyzed.} MSR-VTT \cite{xu2016msr}, VATEX \cite{wang2019vatex}, MSVD \cite{chen2011collecting} and MS-COCO \cite{lin2014microsoft} (for full details, see the supplementary materials). 

All of these datasets collect multiple ground truth descriptions per visual context, and the ground truth descriptions that they do collect are generated by human annotators (via Amazon Mechanical Turk for these datasets). Unfortunately, very large benchmark datasets such as Conceptual Captions \cite{sharma2018conceptual} and HowTo-100M \cite{miech2019howto100m} often contain only a single description per image/video, of questionable quality as the datasets are not annotated by hand. While datasets like S-MiT \cite{monfort2021spoken} contain human-annotated ground truths, they post-process spoken language with automated speech recognition tools, making the dataset difficult to analyze from an n-gram metric angle. Both ActivityNet Captions \cite{krishna2017dense} and YouCook \cite{zhou2018towards} are dense video description datasets that contain high-quality descriptions, however only contain a single ground truth per video. 

Given the datasets, we will contextualize our experiments through the lens of several standard metrics for visual description. The BLEU (or BLEU@N) \cite{papineni2002bleu} score is a measure of n-gram precision, the ROUGE-L \cite{lin2004rouge} score is a measure of longest common sub-sequence recall, the METEOR \cite{banerjee2005meteor} score is a F1-oriented alignment-based metric, and the CIDEr \cite{vedantam2015cider} score is a TF-IDF weighted similarity metric. For more details of the individual metrics, see Aafaq \etal \cite{aafaq2019video}. Recently, metrics which focus more on including visual content directly such as TIGEr \cite{jiang2019tiger} and FAIEr \cite{wang2021faier} have shown improvements in human-judgement correlation and scores such as CLIP-score \cite{radford2021learning}, BERT-score \cite{bert-score}, and SMURF \cite{feinglass2021smurf} have been shown to closer approximate semantic content. While improving the metrics is an extremely important area of research, we also believe that analyzing both why current metrics are failing and what patterns models exploit to optimize these metrics, can give essential insight into model improvements.

We selected a set of recent works from the field as representing the state of the art. For visual description on MSR-VTT and MSVD, we refer to SemSynAN \cite{perez2021improving}, a recent work that uses semantic embeddings based on POS tagging to achieve strong results. SemSynAN was not evaluated on the VATEX dataset, so for VATEX, we refer to the performance of MGRMP (Motion Guided Region Message Passing) \cite{chen2021motion}, a recent method for visual description which leverages message passing between object regions. For MS-COCO, we refer specifically to Vin-VL \cite{zhang2021vinvl}, a method that uses object-level attention and vision and language pre-training for visual description.

%% file: sec/3_motivation.tex
\section{How Can Models Outperform Humans?}
\label{sec:motivation}

Recently, there has been a strong contrast between the metrics-based evaluation of methods for generating visual descriptions on data sets and whether those methods generalize to real-world use cases \cite{stefanini2021show}. The goal of our analysis in this paper is to  understand some of the core reasons why models are failing to generalize and to make recommendations for the future design of datasets, models, and metrics, in an attempt to avoid further generalization shortcomings. 

A core indicator of the difficulty of using standard metrics to improve generalization is that the ``leave-one-out" performance of the ground truths for each dataset is typically poor. Because we investigate datasets that have more than a single ground truth sample per visual context, we can measure the metric scores between a randomly sampled ground truth, and the remaining ground truths for that visual context. When averaged over many trials, this stochastic approach generates an estimate of human performance on the dataset (see the supplementary materials for details). 

Our results are summarized in \autoref{fig:motivation}. We can see that SOTA methods significantly outperform this estimate of human performance on the MSVD, MSR-VTT, and MS-COCO datasets. This result is not only counter-intuitive, but detrimental to progress in the field of video description, as it draws into question the usefulness of standard metrics as an indicator of model performance and generalization. These results motivate questions of understanding: ``Why, and how, do models exploit the current metrics to achieve strong performance?" and ``How can we limit the the exploitation of N-Gram centric metrics".  The goal of the next several sections is to explore these questions through the lens of data diversity. Through analysis of single-token, n-gram, within-sample, and cross-sample diversities, we demonstrate how linguistic patterns affect models and metrics and explore how we can mitigate these effects.

%% file: sec/4_1_h1.tex
\section{Single Sample Diversity}
\label{sec:h1}

To understand and analyze the impact of caption diversity on both model and metric quality, we need to first understand the diversity of the dataset itself. Many datasets use ``vocabulary size", the number of unique tokens (usually words), as a proxy for the diversity of the dataset, however we hypothesize that this metric alone does not tell the full story of token level diversity in visual description datasets. In this section, we analyze the diversity of visual description datasets bottom-up, starting from tokens and working our way up to measures of n-gram complexity.

\subsection{Token-Level Diversity}
\label{sec:tld}


\begin{table}
   \footnotesize
    \centering
    \begin{tabularx}{\linewidth}{Xcccc}
    \toprule
        \textbf{Dataset} & \textbf{Unique} & \textbf{WS-Unique} & \textbf{Head}  \\
    \midrule
       MSVD & 9455 & 11.8\% & 944 \\
        MSR-VTT & 22780 &  21.55\% & 1636 \\
        VATEX & 31364 & 24.87\% & 1363 \\
        MS-COCO & 35341 & 33.76\% & 824 \\
    \bottomrule
    \end{tabularx}
    \caption{Vocabulary metrics for each of the datasets. Unique: The number of unique tokens. WS-Unique: Average percentage of tokens that are unique within a sample (image/video). Head: The number of unique tokens comprising 90\% of the total tokens.}
    \label{tab:vocab_mets}
    \vspace{-0.5em}
\end{table}

\autoref{tab:vocab_mets} provides a token-level analysis of each of the datasets. In addition to reporting the number of unique tokens in the dataset, we also introduce three new measures of diversity: 
Within-Sample uniqueness, which measures the percentage of tokens that are unique within a particular image/video; and ``Vocab-Head", which measures the number of tokens making up 90\% of the tokens in the dataset.
Within-sample diversity ranges between 11\% and 35\%, suggesting that within samples, the descriptions are relatively varied. We discuss the impact of within-sample diversity in \autoref{sec:h3}.
As expected, a small fraction of tokens represent 90\% of the occurrence in most of the datasets. In MS-COCO, 2\% of the tokens represent 90\% of the occurrences, while at the other extreme 10\% of the tokens are required for MSVD. This begs the question: how does the effective vocab size impact performance? 

To validate how effective vocab size impacts performance, we used the same setup as in \autoref{sec:motivation} to compute the performance of the ground truths, however, replaced tokens in the tail of the token distribution with unique ``UNK" tokens. Performance dropped significantly in all cases, with the most dramatic drop for MSVD (drop of 63.87\%) and the least for MS-COCO (drop of 51.23\%). MSR-VTT experienced a decrease of 58.66\% and VATEX experienced a decrease of 56.20\%). Counter-intuitively, the longer the tail, the less performance decreased. This result, confirmed in classification by Tang \etal \cite{tang2020long}, implies that models which generate from a limited vocabulary are advantaged (in terms of n-gram performance) when the head is relatively small, leading to undesirable generation behavior.

Following Wang \etal \cite{wang2019vatex}, we analyze the datasets at the level of the parts of speech in the dataset (See the supplementary materials for details). VATEX has more than 2 verbs per caption on average (by design, see \cite{wang2019vatex}) while the other datasets have at most 1.3 verbs. While VATEX is the most linguistically complex, the distribution has significantly different base statistics, likely explaining poor cross-dataset generalization to VATEX from MSR-VTT and MSVD trained models. MSR-VTT is the most diverse from an object perspective (1512 nouns representing 90\% of the noun mass), which lends additional support to the observations by Zhang \etal \cite{zhang2020object}, who find that a strong object detector and good object features are necessary for strong MSR-VTT performance. Notably, MS-COCO has a very high within-sample noun diversity, suggesting that many of the captions in MS-COCO focus on different objects in each sample, and supporting hypotheses introduced in Anderson \etal \cite{anderson2018bottom} based on multiple-object attention for this dataset.

\subsection{N-Gram-Level Diversity}
\label{sec:ngld}

\begin{table}
    \footnotesize
    \centering
    \begin{tabularx}{\linewidth}{Xccccc}
    \toprule
        \textbf{Dataset} & \textbf{TPC} & \textbf{EVS-2} & \textbf{EVS-3} & \textbf{EVS-4} & \textbf{ED@10} \\
        \midrule
        MSVD & 7.03 & 47.83\%  & 25.29\% & 14.67\% & 2.90 \\
        MSR-VTT & 9.32 & 52.96\% & 26.44\% & 13.68\% & 2.88 \\
        VATEX & 15.29 & 54.84\% & 32.60\% & 18.86\%  & 3.38 \\
        MS-COCO & 11.33 & 53.91 \% & 32.59\% & 20.56\% & 3.51 \\
        WT-103 & 87.04 & 95.19 \% & 34.49\% & 17.81\% & 3.72 \\
    \bottomrule
    \end{tabularx}
    \caption{Effective vocab size (EVS), number of tokens per caption (TPC) and Effective Decision (ED@N). The EVS-n is the percentage of n-grams that do not act like 1-grams in the dataset. A large EVS-n means that language is more diverse, while a small EVS-n means that there are very few combinations of possible n-grams. The ED@N is the expected number of decision that a model has to make when generating captions of length N. WT-103 is WikiText-103 \cite{merity2016pointer}, a common natural language dataset.}
    \label{tab:evs}
    \vspace{-1em}
\end{table}

From tokens, we can move on to exploring how the tokens fit together. One of the major issues in overall dataset diversity is a tendency for language models to accentuate a lack of n-gram diversity, leading to domination of common n-grams over visually likely n-grams \cite{hendricks2018women}. A standard metric reported by Wang \etal \cite{wang2019vatex} in VATEX is the number of unique n-grams in the dataset, however, we find that alone, the number of unique n-grams does not allow for strong comparison between datasets, both because the number is not normalized, and the number of n-grams says little about the overall distribution of those n-grams.

Instead of only looking at the number of n-grams, in order to measure the amount of n-gram diversity that is introduced into a dataset, we introduce the N-Gram Effective Vocab Size metric (EVS-N), which measures the percentage of n-grams that do not act like 1-grams in practice. Formally, EVS-N is the percentage of tokens for which an N-gram language model has zero conditional variance (i.e. the percentage of tokens for which an n-gram language model does not assign 100\% probability to a single next token). This metric can be thought of as a language-generation complexity metric --- a higher EVS means that it will be more difficult for a model to memorize captions, while a low EVS suggests that models need only determine the first few words in order to generate a high-quality caption. \autoref{tab:evs} shows EVS-N performance, and a shocking result. The EVS-2 is approximately 50\% for all datasets, suggesting that in the majority of cases, the model is able to make only one decision to generate two tokens, contrasting with WikiText-103 \cite{merity2016pointer}, where the EVS-2 is 95.19\%. 

In addition to just understanding the EVS, we can combine the EVS scores with the average number of tokens in the dataset to compute the average number of ``decisions" that a model has to make during generation. The ED@N, or expected number of decisions made in a description of length N is also given in \autoref{tab:evs}. Formally, the ED@N is the expected number of tokens in a description of length $N$ for which an n-gram language model of the dataset has non-zero variance conditioned on the sentence so far. Surprisingly, most of the datasets have very similar ED scores (despite their differing average token lengths), and the number is low: only 3-3.5 decisions have to be made on average to get the desired caption. This low number has major implications in the quality of the captions: the fewer the number of decisions that need to be made at training, the less diverse the captions will be during test time, and the less likely models trained on the low-ED data will be able to generalize to fine-grained differences between samples. Further, this means that the number of captions models will be able to generate is restricted to $V^{ED}$, where $V$ is the size of the vocab, a notably smaller number than expected with large vocab sizes, and long captions.  We believe that this is one of the reasons that non-auto-regressive approaches such as those in Liu \etal \cite{liu2021o2na} and Yang \etal \cite{yang2021non} are able to perform so well on these datasets: they can focus on the visual information, and don't have to worry about the syntactic structure as it is similar for all descriptions.

%% file: sec/4_3_h3.tex
\section{Within Sample Diversity}
\label{sec:h3}

While we have seen that token-level diversity is important for the generation of high quality captions, we also want to understand how within-sample diversity (i.e. diversity within a collection of ground truths for a single visual context) impacts the performance of visual description models. 

\begin{figure*}
    \centering
    \includegraphics[width=\linewidth]{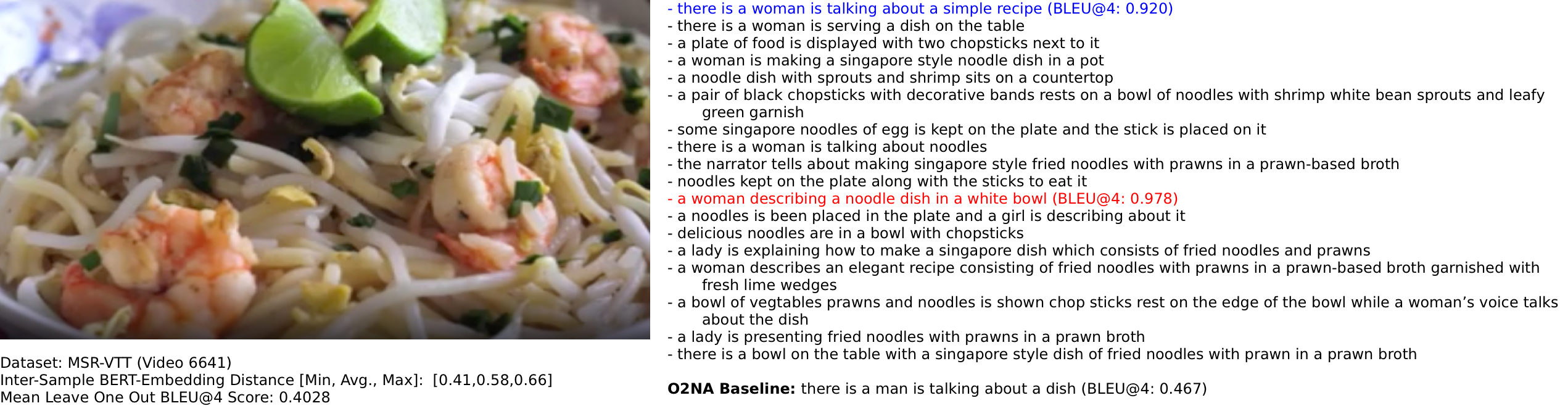}
    \caption{A qualitative example from MSR-VTT demonstrating several diversity effects. The blue description is a description with the minimum distance from the sentence embedding mean, while the red description maximizes the mean BLEU@4 score to all other captions in the sample. Notably, both captions are much more generic than the other captions in the data, a trend which is consistent across all samples. We can see that the variance within this sample is high, however the tokens themselves are similar (annotators select similar tokens for the same sample). Captions are ordered from top to bottom by similarity to the mean caption embedding (See section \ref{sec:h3}). }
    \label{fig:q1}
    \vspace{-0.5em}
\end{figure*}

To define how much within sample diversity there is in a dataset, there are several methods that we can use. One metric, common to many papers, is an analysis of how many captions in each sample are novel. VATEX (100\%) and MS-COCO (99.9\%) have high caption novelty, while MSR-VTT (92.66\%) and MSVD (85.3\%) contain somewhat less exact novelty. Further, we could look at within-sample token diversity (shown in \autoref{tab:vocab_mets}), which suggests that within a sample, diversity is actually relatively high, with 11\% to 33\% of tokens being unique within a sample. Further, the within sample verb (15\% to 56\%) and noun (13\% to 35\%) uniqueness is relatively high as well, suggesting that individually, captions discuss unique parts of a visual context (Full results are given in the supplementary materials). This is demonstrated qualitatively in \autoref{fig:q1}.

\begin{figure}
    \centering
    \includegraphics[width=\linewidth]{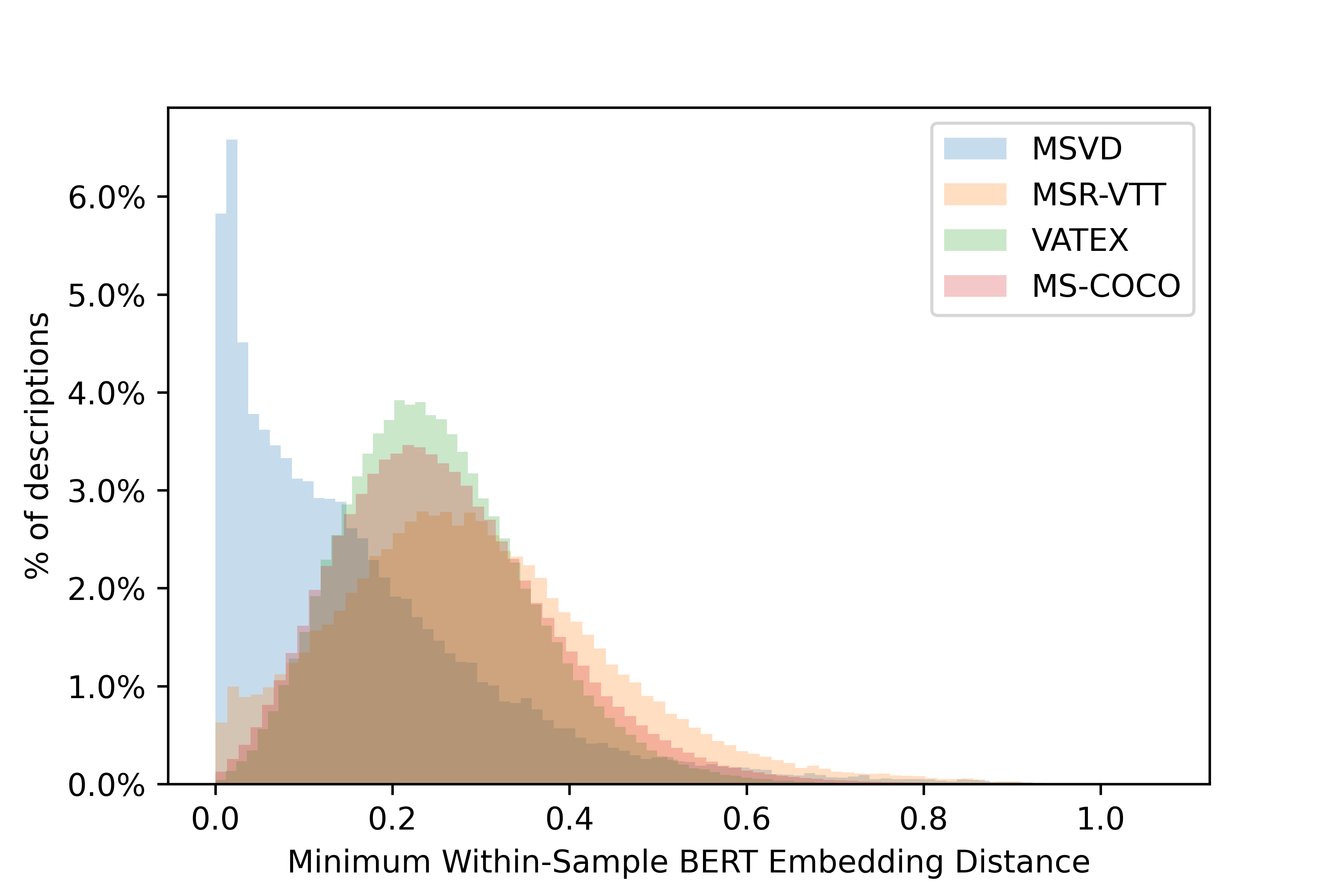}
    \caption{Histogram of within-sample minimum distances under the MP-Net \cite{song2020mpnet} BERT-style embeddings. MSVD and MSR-VTT both have a high number of descriptions which have 0 within-sample minimum distance, while MS-COCO and VATEX have a higher within-sample diversity. }
    \label{fig:within_distance}
    \vspace{-1em}
\end{figure}

The issue with these measures of novelty is that they account only for novelty at the caption or token level by exact matching, but do not directly target the semantic novelty of the captions. In order to look closer at within-sample diversity, we compute the pairwise semantic distance between each description and all other unique descriptions in the sample using the cosine distance between MP-Net embeddings \cite{song2020mpnet} trained for sentence similarity. \autoref{fig:within_distance} shows the minimum of the inter-sample cosine distances, a metric we call sample redundancy. Notably, almost 10\% of the samples in MSVD have a very close semantic match, suggesting that MSVD has more semantically redundant information than other description datasets.

\begin{figure}
    \centering
    \includegraphics[width=\linewidth]{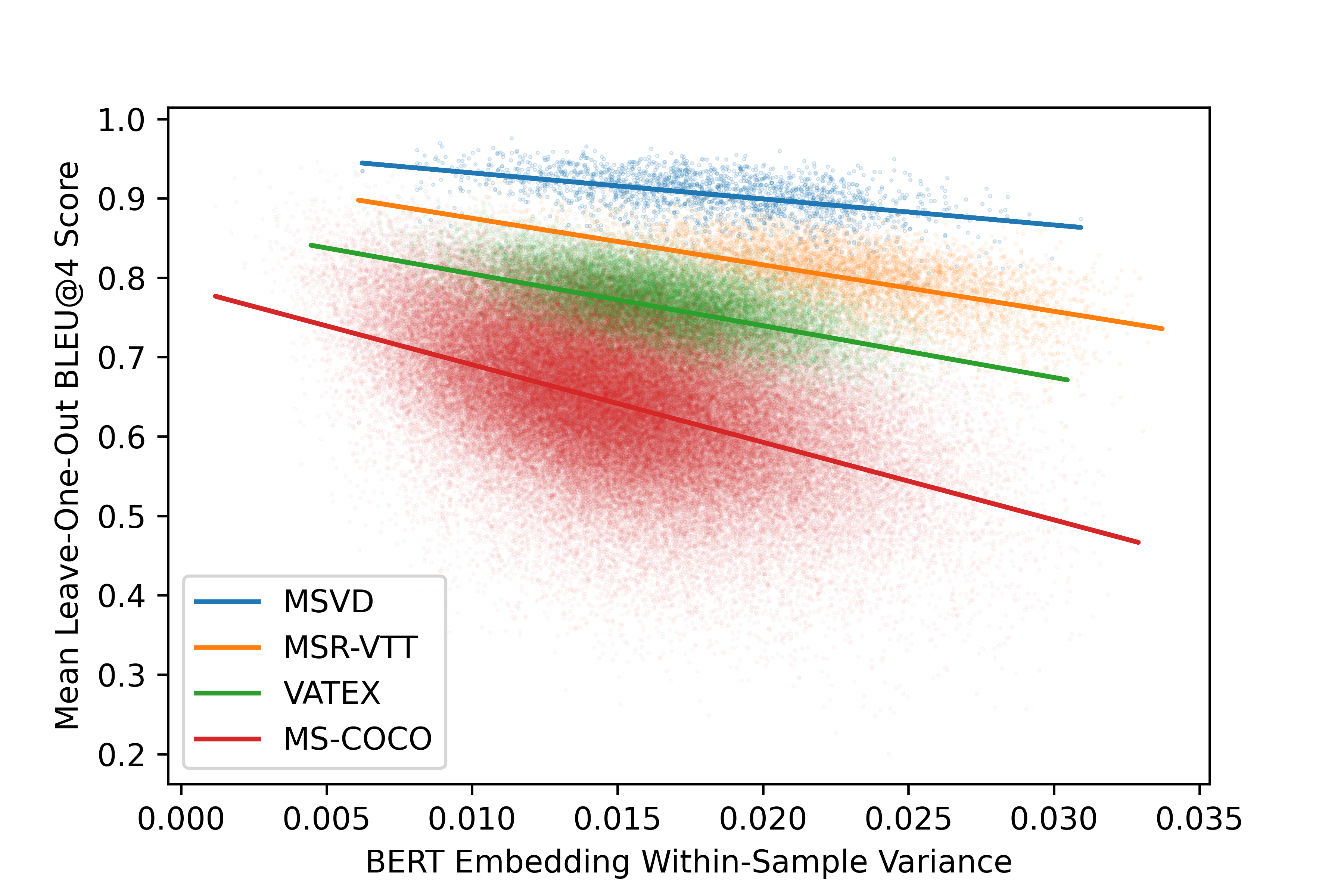}
    \caption{Plot showing the relationship between semantic variance and the performance of leave-one-out ground truth estimates of human performance on the BLEU@4 metrics. As we increase semantic variance, the average minimum distance between ground truth samples increases, and metric performance falls. }
    \label{fig:ws_loo}
    \vspace{-1em}
\end{figure}


Sample redundancy is both a blessing and a curse. Datasets that have very high sample redundancy will tend to have high performance on leave-one-out ground truth metrics, as most of the ground truth captions will share large amounts of information. This means that pair-wise metrics such as the standard n-gram metrics will often perform well, as any generated sample should also lie close to at least one ground truth sample. Unfortunately, as we increase the number of diverse ground truths (increase the sample variance), the minimum distance between samples increases (See the supplementary materials for a figure). Because of this increase in distance, the leave-one-out performance of ground truths decreases, as shown in \autoref{fig:ws_loo}, leading to a breakdown of the n-gram metrics (and all metrics that rely on a single-sample pairwise comparison to the set of ground truths). This effect is what causes SOTA models to outperform leave-one-out samples as demonstrated in \autoref{sec:motivation}. While ideally, metrics should be independent of the variance in the ground truth data, for the datasets we analyze in the paper it is clear the sample variance is sufficient that this is not the case. Interestingly, the leave-one-out fall-off occurs at different rates for the different datasets, suggesting that some datasets are more-redundant to semantic variance than others: while we hypothesize that this is due to the choice of tokens and distribution of semantic structure, it is interesting future work to confirm this hypothesis.

Why are SOTA models immune to the effects of sample variance? It's important to note that when evaluating models, we only look at \textit{a single sample from the model distribution}. We hypothesize that instead of attempting to approximate the full distribution of captions, models are picking up on trends between samples in the data, such as a wealth of descriptions that contain simple semantic structures (as described in \autoref{sec:h1}) or individually strong training descriptions (which we will discuss in \autoref{sec:h2}) which allow the model to reduce the effective variance of the ground truth dataset during the evaluation phase by ignoring most of the ground truth captions, and only focusing on a specific subset of descriptions. While these trends are likely model-specific, we believe it is important future work to quantify and understand the kinds of descriptions that models learn to approximate, and more closely monitor the effects of over-fitting to a small subset of captions to reduce the effects of ground-truth sample variance.

The effect of reducing semantic variance appears in practice via a training trick exploited by both Perez \etal \cite{perez2021improving} and Liu \etal \cite{liu2021o2na} who find that \textit{decreasing} the number of reference captions during training leads to improved evaluation performance on n-gram metrics. By artificially restricting the semantic variance of the training dataset, models are able to over-fit to a smaller subset of semantically redundant captions, and exploit current pairwise metrics. 

Thus, we are stuck in a catch-22 when it comes to adding more captions per sample. If we increase the number of captions, we decrease our metrics' ability to accurately discern caption quality, however if we reduce the number of captions, we can improve the accuracy of current metrics, and obtain models that achieve higher metric scores, at the cost of bland and generic captions.

%% file: sec/4_2_h2.tex
\section{Dataset Level Diversity}
\label{sec:h2}

Not only do sample level diversity and within-sample diversity have important impacts on models and metrics, but dataset-level conceptual diversity matters as well. A common criticism of captioning models is that they are not generative, but instead, reproduce captions from the training set based on a set of global criteria. In general, we hypothesize that a lack of diversity in the dataset, both in the lack of overall visual concept diversity, and the exact distribution of that diversity in the dataset itself leaves models vulnerable to choosing classification over generation. We further hypothesize that a lack of conceptual diversity leads models to produce a few generic captions based on high-level visual features, instead of generating semantically detailed captions. In order to support this hypothesis, we attempt to answer two questions: ``how much performance can we achieve with classification alone?" and ``how much does the explicit selection of visual samples encourage models towards classification over generation?"

\subsection{How many captions make up a dataset?}
\label{sec:coreset}

\begin{figure*}
\centering
\begin{subfigure}{0.31\linewidth}
    \centering
    \includegraphics[width=\linewidth]{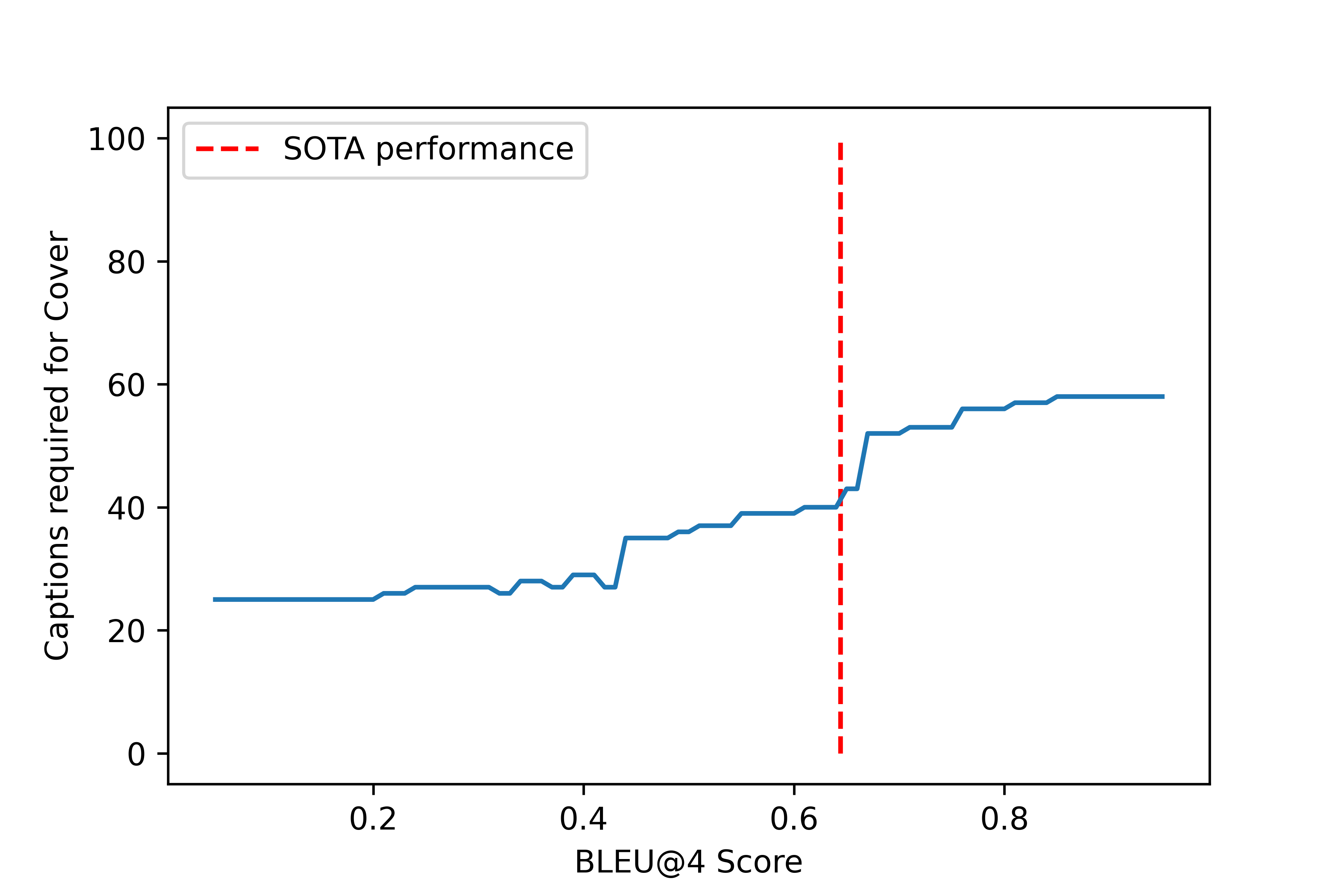}
    \caption{MSVD Dataset}
    \label{fig:msvd_cs}
\end{subfigure}
\hfill
\begin{subfigure}{0.31\linewidth}
    \centering
    \includegraphics[width=\linewidth]{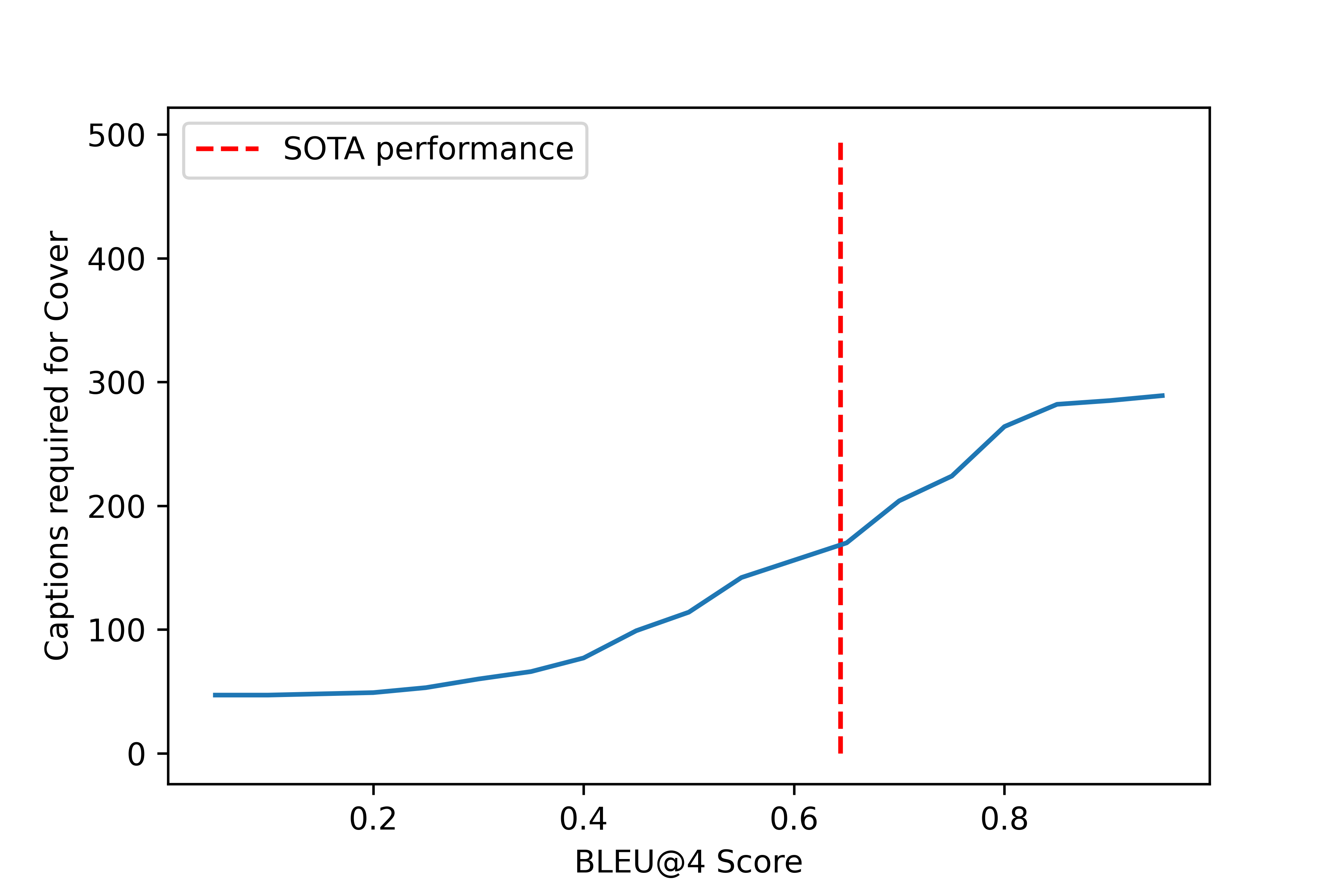}
    \caption{MSR-VTT Dataset}
    \label{fig:msrvtt_cs}
\end{subfigure}
\hfill
\begin{subfigure}{0.31\linewidth}
    \centering
    \includegraphics[width=\linewidth]{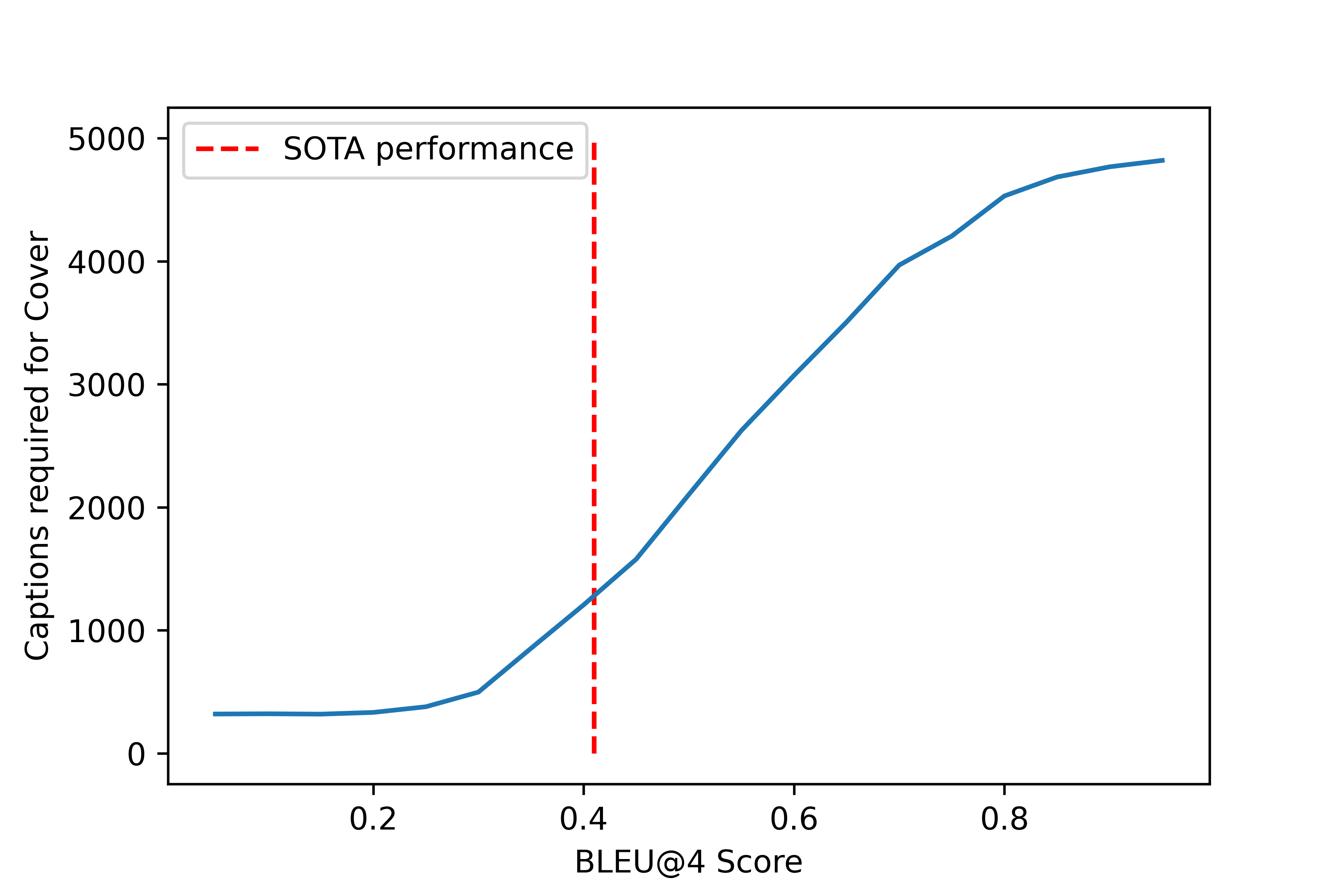}
    \caption{MS-COCO Dataset}
    \label{fig:mscoco_cs}
\end{subfigure}
\caption{For several datasets, how many captions from the training dataset are required to achieve a particular BLEU@4 score on the test set. We can see that in the optimal case, only a few (58 for MSVD, 197 for MSR-VTT, 1578 for MS-COCO) captions are required to achieve SOTA performance on the dataset. Notably, MS-COCO uniquely requires a unique description for each image.}
\label{fig:core_set}
\vspace{-1em}
\end{figure*}

One interesting question to ask is, how many captions do you reasonably need to use in order to solve a dataset to a particular score? This metric is a reasonable proxy for concept-level diversity, and can more globally measure the performance of a model. To answer this question, we used a greedy approximation algorithm for optimal set cover to approximate the minimum number of captions from the training set that need to be chosen for MSR-VTT and MSVD in order to achieve a particular BLEU@4 score on the validation set. We don't compute this number for VATEX/MSCOCO or metrics beyond BLEU due to the computational cost of computing a full matrix of caption distances.  \autoref{fig:core_set} demonstrates the results of this experiment. We can see here that to achieve SOTA BLEU@4 performance, we need only to select optimally from a set of 43 captions in the case of MSVD, and 156 captions in the case of MSR-VTT. Even further, it's interesting to see that with only 58 captions in MSVD and 289 captions in MSR-VTT, we can achieve almost optimal BLEU scores.

This particular result, combined with the fact that models only need to make a few token-level decisions when generating language (See \autoref{sec:ngld}) appears to be a real cause for models producing generic captions. Not only do models not have to make many decisions, but overall, they don't have to select from many visual concepts either.

\subsection{Does the feature set matter?}
\label{sec:fsi}

Caption models are limited not only by a classification effect but also by the concept-level diversity of the feature extractors that they use. When models rely on particular feature extraction methods, we expect pre-initialized features to bias models towards classification over generation, particularly classification among the concepts present in the pre-training data. Recently, Srinivasan \etal \cite{srinivasan2021worst} showed that these biases can compound - so it seems natural to ask the question: how much do we expect biases in our datasets to compound with feature extractor bias? 

\begin{table}
    \footnotesize
    \centering
    \begin{tabularx}{\linewidth}{Xcccc}
    \toprule
        \textbf{Dataset} & \textbf{ImageNet} & \textbf{Kinetics} & \textbf{COCO} & \textbf{Places} \\
    \midrule
        MSVD & 98.27\% & 38.88\%  & 89.03\% & 55.68\%  \\
        MSR-VTT & 68.88\% & 23.51\% & 59.82\% & 46.44\%  \\
        VATEX & 98.60\% & 40.12\% & 76.86\% & 60.55\%  \\
        MS-COCO & 93.22\% & 8.83\% & 91.70\% & 60.49\% \\
    \bottomrule
    \end{tabularx}
    \caption{Percentage of samples in the visual description datasets which contain at least one description that has a sub-string matching a label from the pre-training dataset.}
    \label{tab:overlap}
    \vspace{-0.5em}
\end{table}

\begin{table}
    \footnotesize
    \centering
    \begin{tabularx}{\linewidth}{Xccccc}
    \toprule
        \textbf{Dataset} & \textbf{GT} & \textbf{ImageNet} & \textbf{Kinetics} & \textbf{COCO} & \textbf{Places} \\
    \midrule
        \scriptsize{MSVD} & 0.453 & 0.652  & 0.442 & 0.634 & 0.470 \\
        \scriptsize{MSR-VTT} & 0.210 & 0.678 & 0.467 & 0.650 & 0.521 \\
        \scriptsize{VATEX} & 0.234 & 0.576 & 0.460 & 0.547 & 0.485 \\
        \scriptsize{COCO} & 0.152 & 0.680 & 0.515 & 0.704 & 0.292 \\
    \bottomrule
    \end{tabularx}
    \caption{Performance on BLEU@4 score when using the best core-set ground truth from overlapping categories. Performance remains surprisingly high when using shared captions, implying that models are able to leverage template captions instead of scene understanding. GT: random within-sample leave-one-out ground truth performance.}
    \label{tab:features}
    \vspace{-1em}
\end{table}

In order to measure how much particular datasets are biased towards particular feature extractors, we compute a concept-level ``overlap" between several popular feature datasets \cite{deng2009imagenet, carreira2018short, lin2014microsoft, zhou2017places}, and the visual description datasets. \autoref{tab:overlap} demonstrates the percentage of samples in the visual description datasets which contain at least one description that has a sub-string matching a label from the pre-training dataset. While exact overlap from labels to descriptions may exclude some cases (for example the label "playing baseball" does not overlap with any description which has only the word ``baseball"), we found that fuzzy matching induced significant numbers of false-positives. This metric thus, represents a lower-bound on the overlap (as can be seen in the case of MS-COCO, where only 91\% of the descriptions contain an object from the official label set).

We can see that in datasets except for MSR-VTT, the dataset overlap with ImageNet is relatively high, likely leading to models which achieve performance based solely on the use of ImageNet features, as the classification effect detailed in both \autoref{sec:coreset} and \autoref{sec:ngld} can be exaggerated. Similarly, for datasets besides MSR-VTT, adding object detection features is likely to exaggerate the classification effect, as the model will be pre-disposed to split samples into object-category bins.

To explore exactly how much classification performance can be achieved splitting only along feature extractor boundaries, we generate sets of captions that match (using exact matching) a particular label in the feature extractor pre-training dataset. For each sample, we generate a hypothesis using a randomly sampled caption from the union of the matching concepts and compute the metric score of that hypothesis (See the supplementary materials for a detailed discussion). The results of this experiment are given in \autoref{tab:features}, and we can see that without sufficient conceptual diversity, models can achieve strong performance by segmenting samples among higher-order labels instead of leveraging visual understanding. 


%% file: sec/5_discussion.tex
\section{Recommendations \& Limitations}
\label{sec:discussion}

Our aim in this work is to demonstrate that there are three unique levels of diversity that need to be maintained when collecting a dataset: Token-level diversity, within-sample diversity, and dataset conceptual diversity.

In \autoref{sec:h1} we showed that a lack of token diversity diversity can lead to simple captions from a core data level: few decisions need to be made to generate captions, and a large number of the tokens responsible for this generation are relatively common, opening the door for potential limits to model diversity. Token-level diversity is primarily controlled during the labeling phase of dataset collection, so we believe that both when researchers collect novel data, and when they are building splits for current datasets, they should focus on token diversity. Primarily, to encourage models to generate from a diverse set of captions, we recommend maximizing the ED@N score from \autoref{sec:h1}, along with increasing token EVS by improving the diversity of collected captions. Prompts encouraging crowd-source workers to include higher semantic detail and limits on sentence complexity (such as those introduced in VATEX \cite{wang2019vatex} and Barbosa \etal \cite{barbosa2019rehumanized}) could help prevent token-diversity effects from appearing in downstream models. 

On the other hand, collecting too many ground truths, as discussed in \autoref{sec:h3} presents a model training issue. Currently, models are trained to reduce semantic variance, which can lead to captions which are less complex than we expect. We believe that it is essential future research to explore how to account for the fact that variance in ground truth video descriptions is signal and not noise. Methods for managing multi-modal conditional distributions such as Slade \etal \cite{slade1993bimodal} or multi-label learning such as Tsoumakas \etal \cite{tsoumakas2009learning} may represent step towards such methods. Further, metrics that we use reinforce semantic variance effects by computing maximums with single samples. We believe that investigating metrics which focus on comparing multiple model samples to the full set of ground truth samples represents a possible solution. By forcing models to approximate the entire ground truth distribution we may avoid creating models which optimize away variance in the data. 

Finally in \autoref{sec:h2}, we discussed how a lack of diversity at a concept level can impact the performance of models. When metrics have fewer global concepts, or high overlap with feature extraction methods, they are more likely to trend towards classification over generation. In order to remedy this effect, we recommend the creation of datasets through sampling independent from the label sets of feature models. We additionally recommend that when creating training, validation, and testing splits in the dataset, the concept-level diversity is monitored to avoid introducing potential feature or concept biases with respect to popular feature extraction methods. 

\paragraph{Visual Description Toolkit} Alongside this work, we are releasing a new toolkit\footnote{Toolkit available at \url{https://github.com/CannyLab/vdtk}} for visual description dataset evaluation, which is designed to analyze the performance of models (or ground truths) across the axes explored in this work. We hope that by making tools for evaluating visual description datasets easily accessible, we can encourage the field to deeply investigate the sample diversity in their data and predictions. Further, as part of the analysis toolkit, we are also releasing a set of splits and of the validation and test data for the given datasets, designed to test the performance of models along several of the axes that we discuss in this work, including conceptual labels and caption length among others. We hope that such methods for evaluation can help uncover the deviations of the model from the ground truth data, and paint a more complete picture of our descriptive models beyond n-gram scores.

\paragraph{Limitations} While we have demonstrated how diversity at several levels directly impacts the performance of downstream models, we believe that additional research is required to further understand how the problem of visual description differs from classification and natural language processing. In \autoref{sec:h3}, we use several proxies for caption complexity, however it is not immediately clear that such proxies are good measures for the semantic complexity of a caption. As far as we are aware, no such measure of the ``usefulness" of a caption to a visually impaired user exists, that we can use to evaluate our current caption data. Figure \ref{fig:q1} and the additional qualitative examples in the supplementary material) demonstrate some correlations between caption complexity, and the mean caption, however we believe that deeper analysis is necessary.

Our methods are also limited by the choice of metrics used in this work. Explorations of recent metrics such as FAIer \cite{wang2021faier} may indicate that they alleviate diversity effects by focusing on visual information over textual information, and leveraging pre-trained grounding models. While novel metrics may solve some of the problems, the training effects observed in \autoref{sec:h3} remain common between all models, and the diversity in \autoref{sec:h1} and \autoref{sec:h2} are local to the datasets, and will remain regardless of the metric used.

\section{Background \& Related Work}

This is not the first work to analyze video description data from a dataset and metric perspective, however, we believe that it is the first to focus on how dataset diversity and metric choices directly affect caption generalization. Hendricks \etal \cite{hendricks2018women}, Bhargava \etal \cite{bhargava2019exposing}, Tang \etal \cite{tang2021mitigating} and Zhao \etal \cite{zhao2021understanding} have all demonstrated that visual description data is often biased with respect to protected attributes (such as race, gender or religion), and introduced new methods for handling specific biases - however, they do not discuss the impact of general biases on model performance.  Both Smeaton \etal \cite{smeaton2019exploring} and Yang \etal \cite{yang2021deconfounded} demonstrate poor cross-dataset generalization in visual description, and demonstrate that the choice of dataset directly affects model generalization ability, as well as introduce additional model-centric methods for mitigating the impact of dataset effects. These works complement our own, and they support our core hypotheses that we discuss in \autoref{sec:discussion}.

Outside of visual description, the evaluation of how linguistic data and metrics affects the performance of downstream vision and language models is prevalent. Cadene \etal \cite{cadene2019rubi} demonstrate unimodal language biases in visual question answering and Choi \etal \cite{choi2019can} do the same for action recognition. While many papers \cite{yang2020towards, shah2019predictive, li2019repair, singh2020don, clark2020learning, joo2020gender} make recommendations for reducing linguistic bias based on the modeling framework, these works do not focus on the quality of generation, and instead, focus on the equally important trend of models relying heavily on language priors to solve tasks. Barbosa \etal \cite{barbosa2019rehumanized} introduce methods for dataset collection which attempt to reduce linguistic bias, which represents a great leap forward from standard Amazon Mechanical Turk (AMT) collection methods, but does not discuss how the diversity impacts the performance of downstream models beyond balancing language priors. 

%% file: sec/6_conclusions.tex
\section{Conclusion}

In this work we have taken a close look at linguistic diversity in common visual description datasets, and detailed how diversity can impact models and metrics. At the token level, we showed that a lack of diversity impacts the ability of metrics to assess the quality of captions, and the ability of models to generate diverse descriptions. At the sample level, we demonstrated that high within-sample diversity is both a blessing and a curse, leaving us with either a failure of metrics to correctly measure performance, or leaving us with correct metrics, but bland and generic captions. Finally, at the dataset level, we demonstrated that even when single sample and within-sample diversity is maintained, a lack of conceptual diversity at the dataset level can bias models towards visual classification over language generation, opening the door for models which can use a few, generic, samples to solve the visual description task instead of generating captions which are rich in semantics. 

While this work demonstrates the potential pitfalls of a lack of diversity in visual description datasets, we believe that by introducing new tools for analysis, and additional recommendations for data collection and model evaluation, the field will be able to investigate the sources of poor model generalization more closely, and build models which are both robust to visual diversity and can generate diverse, high quality, and semantically meaningful captions.

%% file: sec/X_supplementary.tex
\clearpage
\pagebreak
\appendix
\setcounter{page}{1}
\setcounter{section}{1}
\setcounter{table}{0}
\setcounter{figure}{0}
\renewcommand{\thepage}{S\arabic{page}} 
\renewcommand{\thesection}{S\arabic{section}}
\renewcommand{\thetable}{S\arabic{table}}  
\renewcommand{\thefigure}{S\arabic{figure}}

\twocolumn[
\centering
\Large
\textbf{What's in a Caption? Visual Description Linguistic Patterns and Their Effects on  Models and Metrics} \\
\vspace{0.5em}Supplementary Material \\
\vspace{1.0em}
] 

\appendix
\section{Datasets}
\label{sec:Datasets}

We investigate four primary datasets in this work. An overview of the datasets is given in \autoref{tab:datasets}.

\paragraph{MSR-VTT} The MSR-VTT (MSR Video to Text) \citesupp{s-xu2016msr} dataset is a medium-scale open domain benchmark for visual description. It was originally collected using 257 YouTube search queries across 20 categories, with 118 videos collected for each query (41.2 Hours). The dataset is annotated with 20 captions per video by 1,327 Amazon Mechanical Turk workers. Each video has a duration between 10 and 30 seconds, with an average of two shots per clip. 

\paragraph{VATEX} The VATEX dataset \citesupp{s-wang2019vatex} is a medium-scale open domain video description benchmark, based on a subset of the Kinetics-600 dataset for action recognition. VATEX consists of 41,269 video clips, and each clip is annotated with 10 unique descriptive captions by 2,159 Amazon Mechanical Turk workers. 

\paragraph{MSVD} The MSVD (Microsoft Video Description) dataset \citesupp{s-chen2011collecting} is a small-scale open domain benchmark for video description comprised of 1,970 YouTube clips of 4-10 seconds each, collected by asking Amazon Mechanical Turk workers to link a video, start time, and end time from YouTube that depicts a specific, short action. Each video is then annotated with an average of 41 ground truth descriptions by 835 Amazon Mechanical Turk workers.

\paragraph{MSCOCO} The Microsoft Common Objects in Context (MS-COCO) \citesupp{s-lin2014microsoft} dataset is a large-scale open-domain benchmark for image description. MS-COCO consists of more than 120,000 images of complex scenes including people, animals, and common objects. Each image is annotated with five ground truth descriptions. \\

\begin{table*}
    \centering
    \footnotesize
    \begin{tabularx}{\linewidth}{Xccccccc}
    \toprule
        \textbf{Dataset} & \textbf{Domain} & \textbf{Categories} & \textbf{Videos} & \textbf{Avg. Length}  & \textbf{Length (hrs)} & \textbf{Annotations / Video}& \textbf{Annotation Method}  \\
    \midrule
        MSR-VTT & open & 20 & 10K & 20s & 41.2 & 20 & AMT \\
        VATEX & open & 600 & 42K & - & - & 10 & AMT \\
        MSVD & open & 218 & 1970 & 10s & 41 & 35.5 & AMT \\
    \midrule
        MS-COCO & open & - & 120K & - & - & 5 & AMT \\
    \bottomrule
    \end{tabularx}
    \caption{An overview of the datasets that we analyze in this paper. All of the datasets are open-domain, with a focus on video description. Additionally, each of the datasets include more than one ground truth description per video, which we use to validate the performance of ground truth data, without collecting additional human results. Notably, all of these methods use AMT as their annotation method.}
    \label{tab:datasets}
\end{table*}

\section{Experimental Details}
\label{sec:exp}

In this section, we present detailed experimental details corresponding to our experiments. Along with these experimental details, we make the code for our work available at \url{https://github.com/CannyLab/vdtk}. Note that numbers may differ slightly between the released code, and our presented experiments due to the tokenization scheme. For our released code, we use the Spacy\footnote{\url{https://spacy.io/}} tokenizer to compute all metrics, as it is significantly more efficient in practice than the Stanford tokenizer\footnote{\url{https://nlp.stanford.edu/software/tokenizer.html}}, however for academic purposes, we compute the metrics with the Stanford tokenizer to avoid tokenization shift. In most cases, the difference in the metrics between tokenization methods is negligible (or very small).

\subsection{Motivation: Leave One Out Ground Truth Performance}
\label{sec:loo}

To generate an estimate of human performance on the selected datasets, we use a procedure called ``leave one out" performance. Let a dataset $\mathcal{D}$ be composed of $N$ samples $S_0 \dots S_N$. For each sample $S_i$, there may be $K_i$ possible reference captions, $C^i_0 \dots C^i_{K_i}$. In order to compute the leave one out performance of human samples for the dataset, we first select a hypothesis caption $H_i \in \{C^i_0 \dots C^i_{K_i}\}$. We then compute the updated reference set $R_i =  \{C^i_0 \dots C^i_{K_i}\} / \{H_i\}$. In the case that $H_i$ is duplicated within $R_i$, \textit{we allow the duplicate to remain} to maximize the possible human score. In the case that there is only one (or fewer) captions for a video, we drop those captions from the computation. We then use the reference sets $R_0 \dots R_N$ and hypotheses $H_0 \dots H_N$ to compute the ``leave-one-out" score for the dataset. Clearly, this is an estimate of the ground truth performance, as it is a random sample of the possible ``leave-one-out" hypotheses sets.

Because some of the metrics (particularly CIDEr) are dataset dependent, it would be intractable to compute all possible hypotheses sets. Instead of computing all possible hypotheses sets, we perform 750 iterations of this sampling procedure and use the mean of the iterations to achieve our final ``leave-one-out" estimates presented in the paper. We found empirically that 750 iterations were sufficient across all of the datasets to achieve a stable mean. The raw values of the ``leave-one-out" estimates are presented in \autoref{tab:loo_raw}, alongside the state of the art results.

\begin{table}
    \footnotesize
    \centering
    \begin{tabularx}{\linewidth}{Xcccc}
    \toprule
        \textbf{Dataset} & \textbf{BLEU@4} & \textbf{METEOR} & \textbf{ROUGE} & \textbf{CIDEr} \\
    \midrule
        MSVD & 0.453 \tiny{(0.644)} & 0.370 \tiny{(0.419)}  & 0.689 \tiny{(0.795)} & 1.038 \tiny{(1.115)}  \\
        MSR-VTT & 0.209 \tiny{(0.472)} & 0.247 \tiny{(0.312)}  & 0.487 \tiny{(0.648)} & 0.426 \tiny{(0.600)}  \\
        VATEX & 0.234 \tiny{(0.342)} & 0.249 \tiny{(0.235)}  & 0.478 \tiny{(0.503)} & 0.611 \tiny{(0.576)}  \\
        MS-COCO & 0.152 \tiny{(0.410)} & 0.228 \tiny{(0.311)}  & 0.438 \tiny{(0.609)} & 0.788 \tiny{(1.409)}  \\
    \bottomrule
    \end{tabularx}
    \caption{Raw leave-one-out score estimates for each of the datasets (SOTA in parentheses). }
    \label{tab:loo_raw}
\end{table}

\subsection{Motivation: Semantically Masked Leave One Out performance}

To test the performance of ground truths without semantic information, we devised an experiment based on the leave-one-out experiments above, however, focused on removing semantic information. To compute this value, we select hypotheses as in \autoref{sec:loo}, however for both the captions in the reference and the captions in the ground truth, we replace any token identified by the Spacy part of speech analysis as a noun, proper noun, or verb with a \textit{unique} mask token. This means that this unique mask token will achieve a 0 in any associated token-based metric, as it will not match any semantic token in the ground truth. Table \autoref{tab:msk_raw} gives the full performance on each of the datasets in the masked setup. 

\begin{table}
    \footnotesize
    \centering
    \begin{tabularx}{\linewidth}{Xcccc}
    \toprule
        \textbf{Dataset} & \textbf{BLEU@4} & \textbf{METEOR} & \textbf{ROUGE} & \textbf{CIDEr} \\
    \midrule
        MSVD & 0.289 \tiny{(0.453)} & 0.097 \tiny{(0.370)}  & 0.442 \tiny{(0.689)} & 0.502 \tiny{(1.038)}  \\
        MSR-VTT & 0.123 \tiny{(0.209)} & 0.085 \tiny{(0.247)}  & 0.387 \tiny{(0.487)} & 0.327 \tiny{(0.426)}  \\
        VATEX & 0.132 \tiny{(0.234)} & 0.201 \tiny{(0.249)}  & 0.391 \tiny{(0.478)} & 0.511 \tiny{(0.611)}  \\
        MS-COCO & 0.079 \tiny{(0.152)} & 0.198 \tiny{(0.228)}  & 0.396 \tiny{(0.438)} & 0.684 \tiny{(0.788)}  \\
    \bottomrule
    \end{tabularx}
    \caption{Raw leave-one-out score estimates under semantic masking for each of the datasets (Non-masked in parentheses). }
    \label{tab:msk_raw}
\end{table}

\subsection{Caption Diversity: Token Metrics}

In this work, we compute several metrics based on token-level diversity, demonstrated in \autoref{tab:vocab_mets} from the main paper. The number of unique tokens is equal to the number of tokens in the dataset as computed by the Stanford PTB tokenizer. This number does not do any lemmatizing or stemming, thus, is an upper bound for the vocabulary complexity. We then compute three additional metrics, the within-sample uniqueness, the between-sample uniqueness, and the 90\% head of the vocabulary. The within-sample uniqueness corresponds to the percentage of tokens that are unique within a sample - i.e. the percentage of tokens that appear exactly once among the references for any particular image or video. We then average this number over all of the samples to get the number presented in \autoref{tab:vocab_mets}. The between-sample uniqueness is a measure of the percentage of tokens in each sample that are unique at the \textit{dataset} level, i.e. the percentage of tokens among the tokens in the reference set of a single sample that do not appear in any other caption in the dataset. These per-sample numbers are then averaged across the dataset to get the number presented in \autoref{tab:vm_complete}. Finally, the 90\% head corresponds to the number of tokens that make up 90\% of the mass of the total number of tokens in the dataset. This is an approximate measure of how long-tailed the distribution is. The 90\% number is selected empirically (further analysis could look at the full cumulative distribution of the token counts). \autoref{tab:vm_complete} replicates \autoref{tab:vocab_mets} from the main paper, however includes between-sample token uniqueness. 

We also compute many of the same metrics restricted to counting nouns and verbs (as identified by the Spacy POS tagger). Each of the above metrics is computed the same way, however instead of considering all tokens, we consider only tokens that are tagged as either nouns or verbs during the computation of the metrics. \autoref{tab:pos_distribution} demonstrates the full results of this experiment, plus an additional metric: the average number of tokens per caption which also appears in \autoref{tab:evs} in the main paper. 

\begin{table}
  \footnotesize
    \centering
    \begin{tabularx}{\linewidth}{Xcccc}
    \toprule
        \textbf{Dataset} & \textbf{Unique} & \textbf{BS-Unique} & \textbf{WS-Unique} & \textbf{Head}  \\
    \midrule
      MSVD & 9455 & 1.21\%  & 11.8\% & 944 \\
        MSR-VTT & 22780 & 0.76\% & 21.55\% & 1636 \\
        VATEX & 31364 & 0.33 \% & 24.87\% & 1363 \\
        MS-COCO & 35341 & 0.22\% & 33.76\% & 824 \\
    \bottomrule
    \end{tabularx}
    \caption{Vocabulary metrics for each of the datasets. Unique: The number of unique tokens. BS-Unique: Average percent of tokens per description that are unique. WS-Unique: Average percent of of tokens that are unique within a sample. Head: The number of unique tokens comprising 90\% of the total tokens.}
    \label{tab:vm_complete}
\end{table}

\begin{table*}
    \footnotesize
    \centering
    \begin{tabularx}{\linewidth}{Xccccccccccc}
    \toprule
        \textbf{Dataset} & \textbf{WSNU} & \textbf{BSNU} & \textbf{WSVU} & \textbf{BSVU}  & \textbf{NC} & \textbf{VC} & \textbf{NH} & \textbf{VH} &  \textbf{NPC} & \textbf{VPC} & \textbf{TPC}  \\
    \midrule
        MSVD & 12.6\% & 1.9\% & 14.8\% & 1.5\% & 4985 & 1773 & 755 & 229 & 2.39 & 1.10 & 7.03 \\
        MSR-VTT & 23.1\% & 1.2\% & 29.4\% & 0.8\% & 12697 & 3639 & 1512 & 293 & 3.28 & 1.32 & 9.32   \\
        VATEX & 26.9\% & 0.67\% & 35.7\% & 0.3\% & 16670 & 4975 & 1161 & 338 & 4.37 & 2.10 & 15.29 \\
        MS-COCO & 34.9\% & 0.41\% & 55.8\% & 0.2\% & 20155 & 4200 & 723 & 184 & 3.71 & 1.02 & 11.33 \\
    \bottomrule
    \end{tabularx}
    \caption{Part of speech distributions for each of the datasets. DS: Dataset. WSNU: Within sample noun uniqueness. BSNU: Between sample noun uniqueness. WSVU: Within sample verb uniqueness. BSVU: Between sample verb uniqueness. NC: Unique noun count. VC: Unique verb count. NH: Noun head (90\% of mass). V: Verb Head (90\% of mass). VPC: Average number of verbs per caption. NPC: Average number of nouns per caption. TPC: Average number of tokens per caption. }
    \label{tab:pos_distribution}
\end{table*}

\subsection{Caption Diversity: N-Gram Metrics}

To explore the diversity of samples at an n-gram level, we introduce two novel metrics, the Expected Vocab Size @ N (EVS@N), and the Expected Number of Decisions @ N (ED@N). Both of these metrics measure the diversity of the language at an n-gram level by exploring the properties of an n-gram language model trained on the dataset. In this section, we discuss the explicit definition of these metrics. For all n-grams, we use an n-gram language model based on tokens extracted with the Stanford PTB tokenizer. In all cases, we pad the references with $[BOS]$ and $[EOS]$ tokens to allow the model to handle the beginning and end of the sequences. For WikiText-103, we create individual reference sentences by splitting on `.` tokens, and pad each of these references individually with $[BOS]$ and $[EOS]$ tokens.

\subsubsection{Expected Vocab Size @ N}

The EVS@N metric is a measure of how many n-grams \textit{do not} act as 1-grams in practice in the dataset. This measure is computed by looking at the entropy of the next-token distribution of an n-gram language model. For a sequence of words $w_0, \dots w_{n-1}$, we first compute the distribution $P(w_n | w_0, \dots, w_{n-1})$. If this distribution has 0 entropy (i.e. it assigns all of the probability mass to a single next token), then we consider this n-gram a ``static n-gram". If the entropy is non-zero, then we consider it a ``dynamic n-gram". The EVS@N can then be computed as the proportion of dynamic n-grams $$\text{EVS@N} = \frac{|\text{dynamic n-grams}|}{|\text{static n-grams}| + |\text{dynamic n-grams}|}$$ This measures a set of effective n-grams in the data (i.e. the size of the n-gram vocab), as it coalesces n-grams where no decisions are made into a single logical unit. 

\subsubsection{Expected Decisions @ N}

The ED@N metric is a measure of how many decisions an n-gram language model has to make for a sequence of $N$ tokens. ED@N is a counting measure of the EVS@N - i.e. how many dynamic n-grams are expected in a sequence of length $n$. For a $K-gram$ language model, this measure is explicitly computed as: $$ \text{ED@N} = 1 + \sum_{i=1}^{N-1} (1-EVS@K)(0) + (EVS@K)(1) $$ In this work, for the first token we use a 2-gram language model ($K=2$), for the second token we use a 3-gram language model $(K=3)$, and for any additional tokens, we use a 4-gram language model $(K=4)$.

\subsection{Sample Diversity: Within Sample Diversity}

We use several techniques to measure the within-sample semantic diversity of the data. In all of these cases, the notion of semantics is somewhat subjective. In this work, we use a BERT-style embedding trained for sentence similarity, called MP-Net \citesupp{s-song2020mpnet} to embed each reference description as a 384-dimensional vector. We leverage the implementation in Sentence Transformers\footnote{\url{https://huggingface.co/sentence-transformers/all-mpnet-base-v2}}, which is pre-trained on over 1 billion sentence pairs. 

\autoref{fig:within_distance} measures the minimum within-sample distances, i.e. it looks for the closest pair of references in each sample, and plots the distance between them. Thus, for a dataset of length $N$ with a set of samples $S_0 \dots S_N$ and captions $S_i^0 \dots S_i^{K_i}$, this histogram plots the distribution over all descriptions of $$H_{ij} = \min_{k \neq n} ||S_i^k - S_i^j||$$ In order to avoid obvious issues with repetition in the semantics, we use only the unique set of captions in a sample, as opposed to allowing for duplicates, which would force $H_i$ to zero for any sample with repeated captions (actually exaggerating the effect in \autoref{fig:within_distance}. We don't allow this in order to avoid biasing our experiments to datasets such as VATEX, which explicitly remove exact duplicates. Close duplicates are not affected, as can clearly be seen by MSVD, which contains a lot of semantic redundancy. Note that this is a distribution over all references (as opposed to samples). 

\begin{figure}
    \centering
    \includegraphics[width=\linewidth]{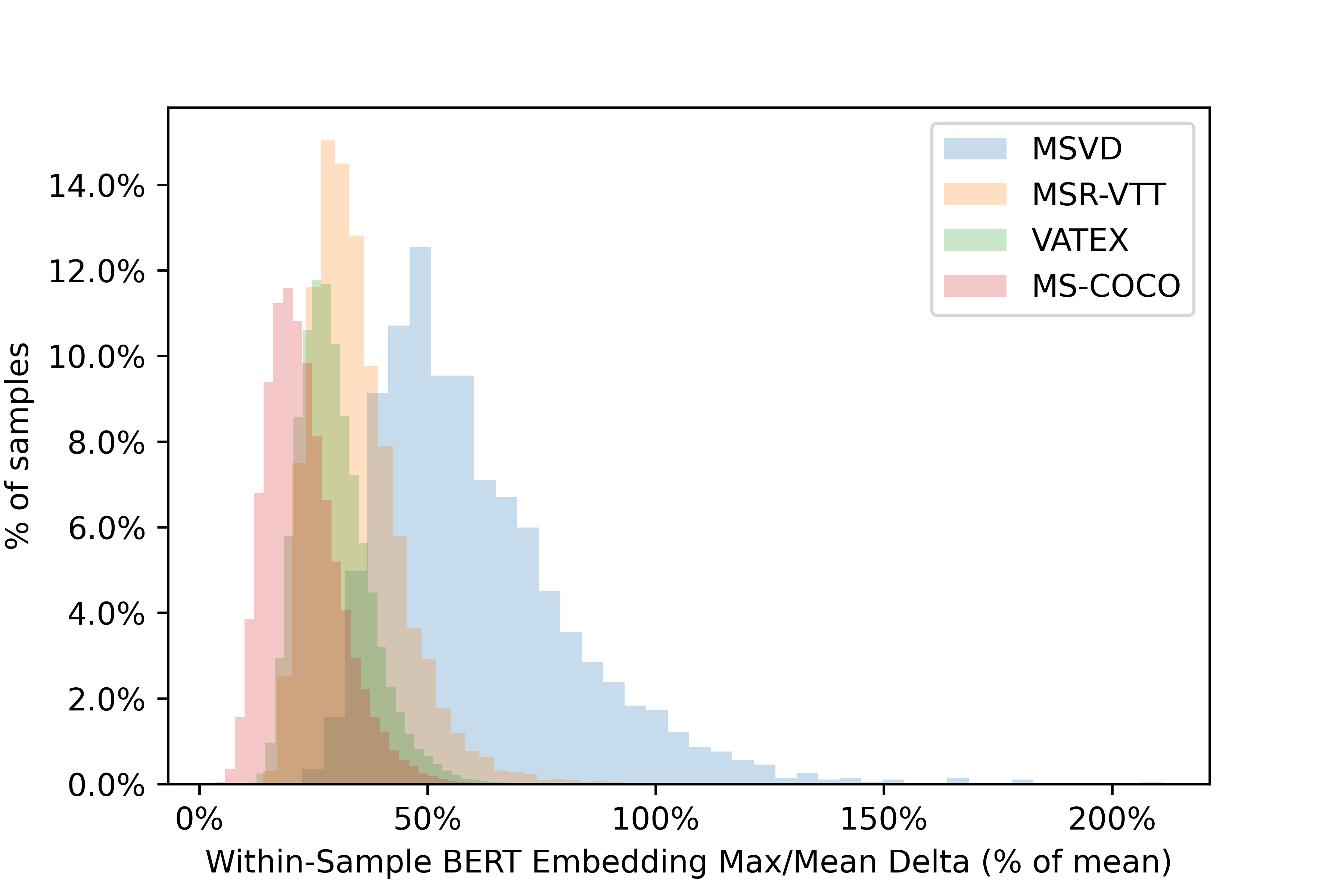}
    \caption{Plot demonstrates the difference between the closest semantic vector, and the mean of the semantic vectors. In all cases, the mean will always be further than the closest sample, however, a low delta suggests a more equal spread of references, while a high delta represents highly redundant samples.}
    \label{fig:ws_delta_bert}
\end{figure}

Another method of measuring semantic diversity is by looking at the spread of the semantics in the sample. While we use the literal variance of the within-sample pairwise distance distribution in \autoref{fig:ws_loo}, we can also look at other measures of spread. \autoref{fig:ws_delta_bert} demonstrates the difference (as a percent of the mean) between the mean of the inter-sample distances and the closest inter-sample distance. When this percentage is high, the descriptions are relatively spread out for a sample, with clusters of descriptions that are close together in semantic space. If the percentage is low, the descriptions for a sample are well-distributed (mostly equidistant) in the semantic space.

\begin{figure}
    \centering
    \includegraphics[width=\linewidth]{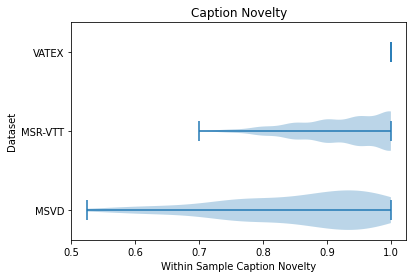}
    \caption{Violin plot demonstrating the distribution of caption novelty - i.e. how many captions in each sample are not exact matches in the text space. As we can see, while the vast majority of captions are novel in some datasets, in datasets like MSVD, there some samples which have high \textit{exact} redundancy. }
    \label{fig:cap_novel}
\end{figure}

\autoref{fig:cap_novel} gives a general overview for the video description datasets of the exact-duplicate distribution of the descriptions. While most of the samples have high within-sample uniqueness, there are some samples that are highly redundant (and in the case of MSVD, have exact-redundancy of as much as $\sim$ 50\%.  
 
\subsection{Dataset Diversity: Number of Ground Truths}

\begin{figure*}
\centering
\begin{subfigure}{0.45\linewidth}
    \centering
    \includegraphics[width=\linewidth]{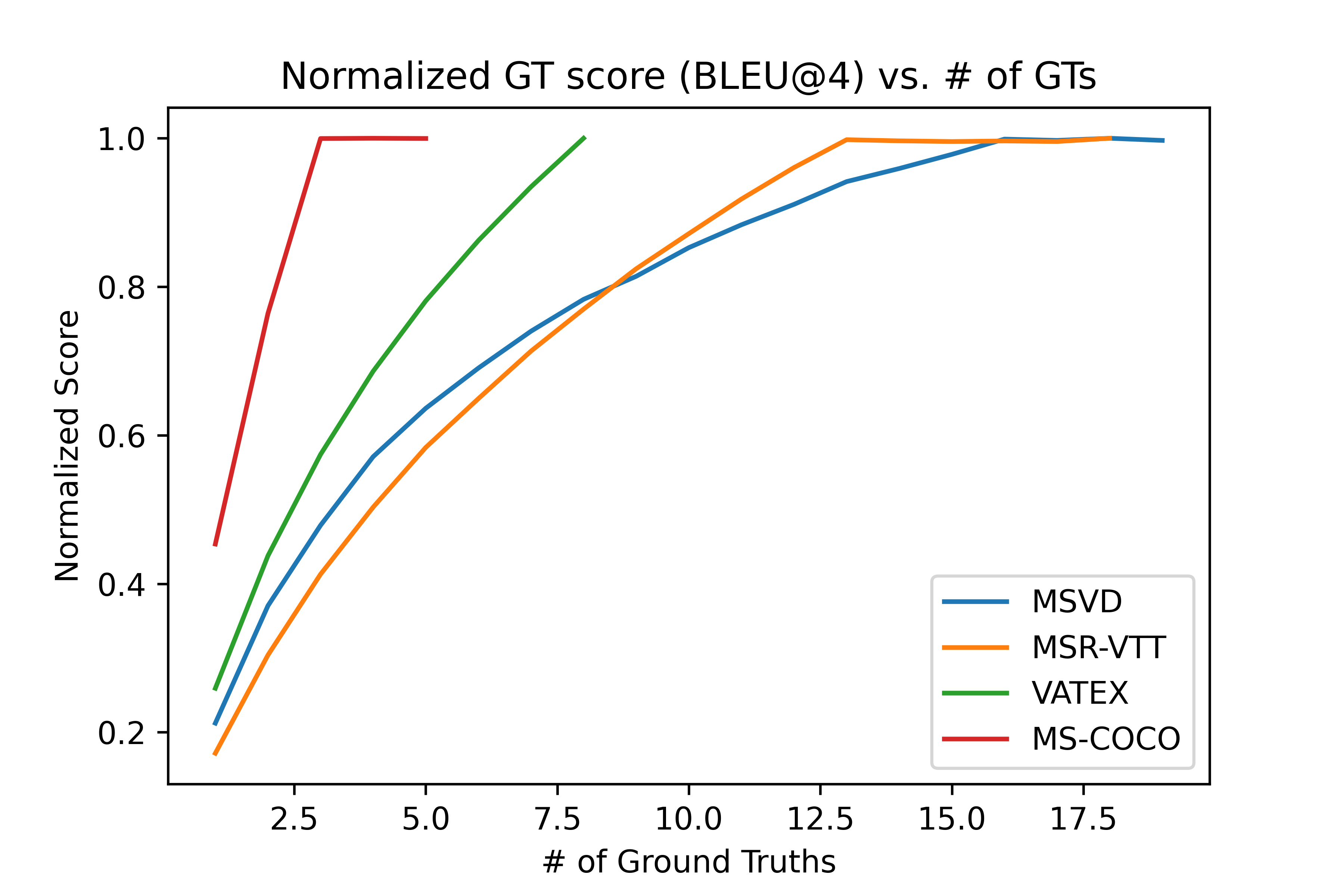}
    \label{fig:ngt_b}
\end{subfigure}
\hfill
\begin{subfigure}{0.45\linewidth}
    \centering
    \includegraphics[width=\linewidth]{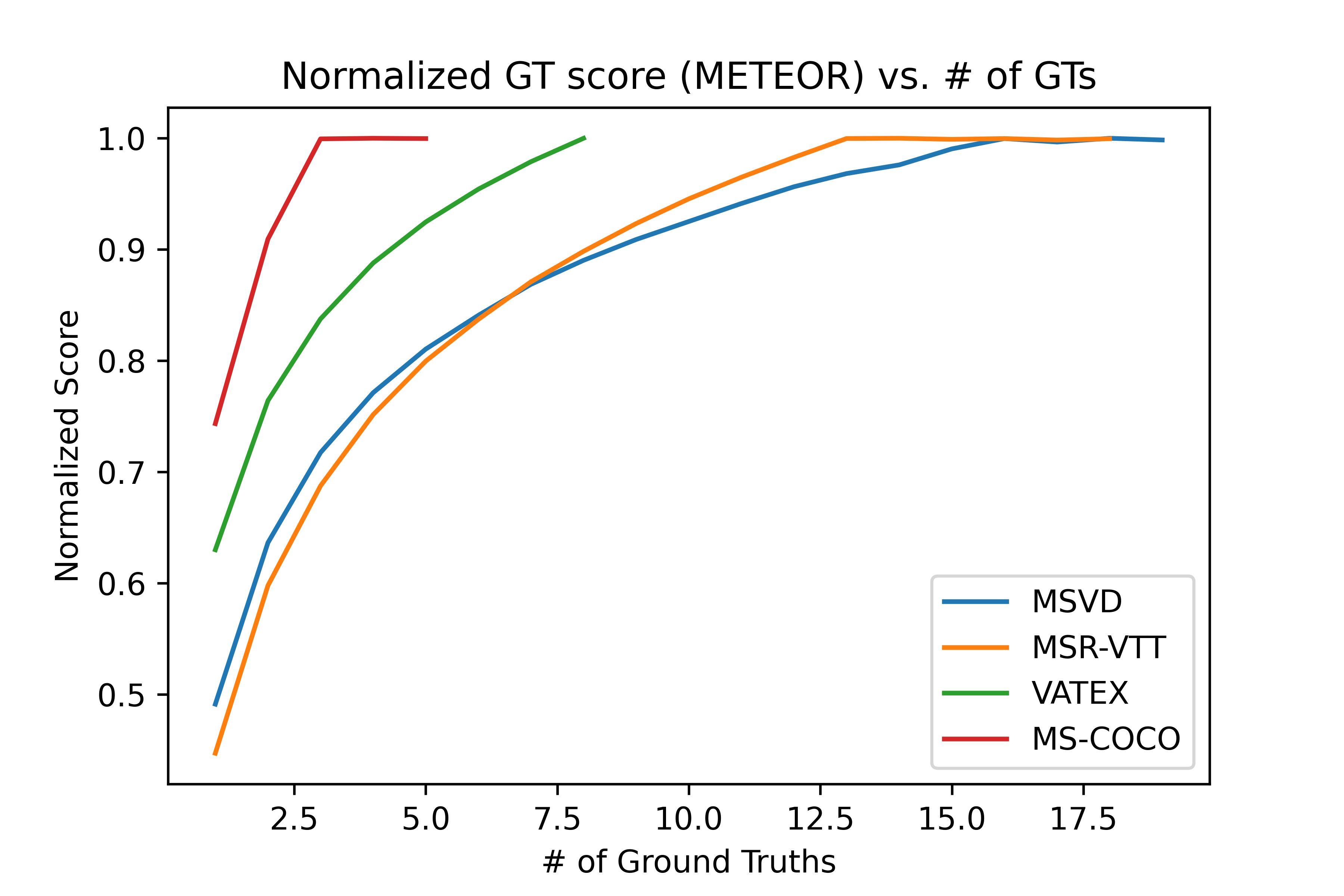}
    \label{fig:ngt_m}
\end{subfigure}
\\
\begin{subfigure}{0.45\linewidth}
    \centering
    \includegraphics[width=\linewidth]{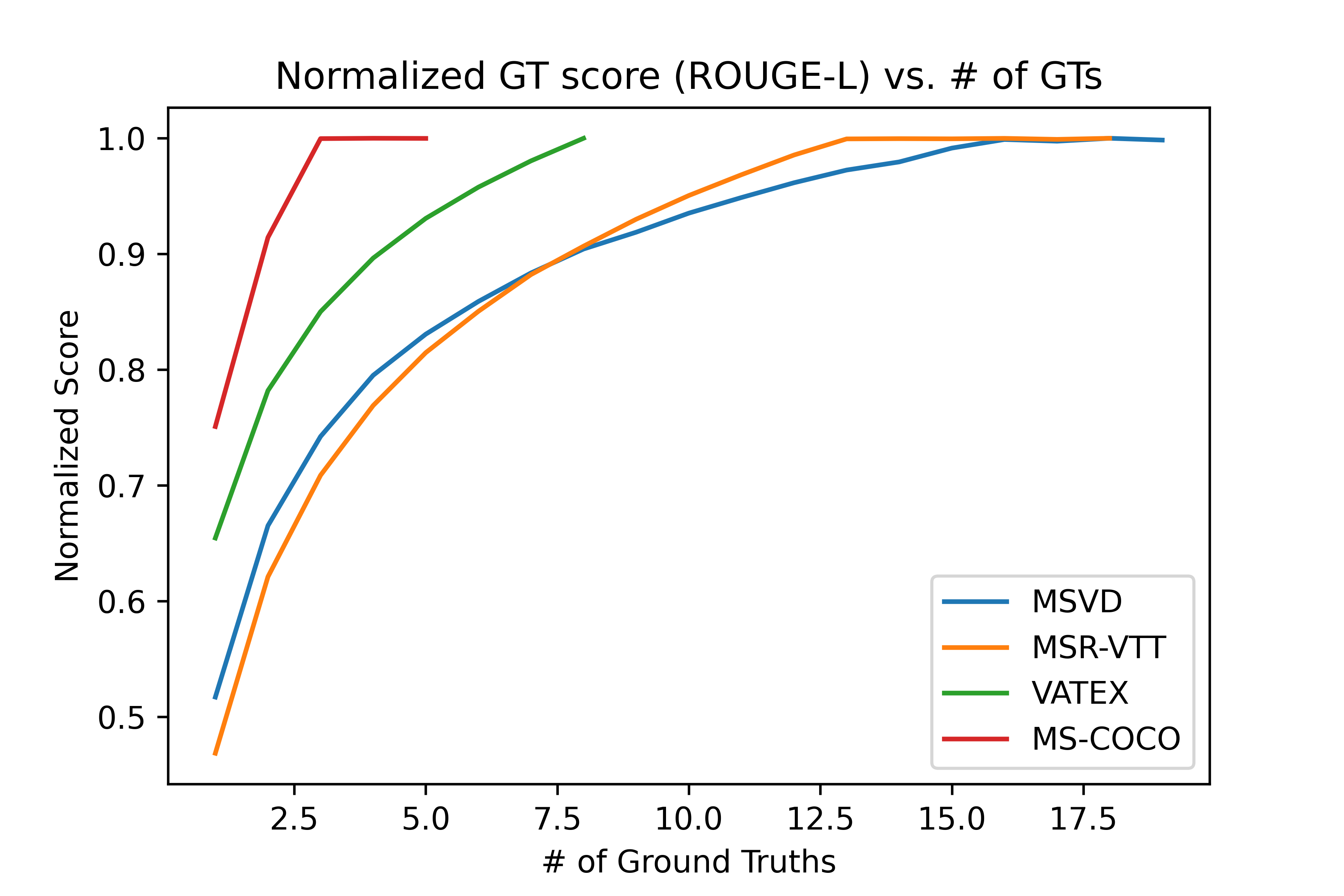}
    \label{fig:ngt_r}
\end{subfigure}
\hfill
\begin{subfigure}{0.45\linewidth}
    \centering
    \includegraphics[width=\linewidth]{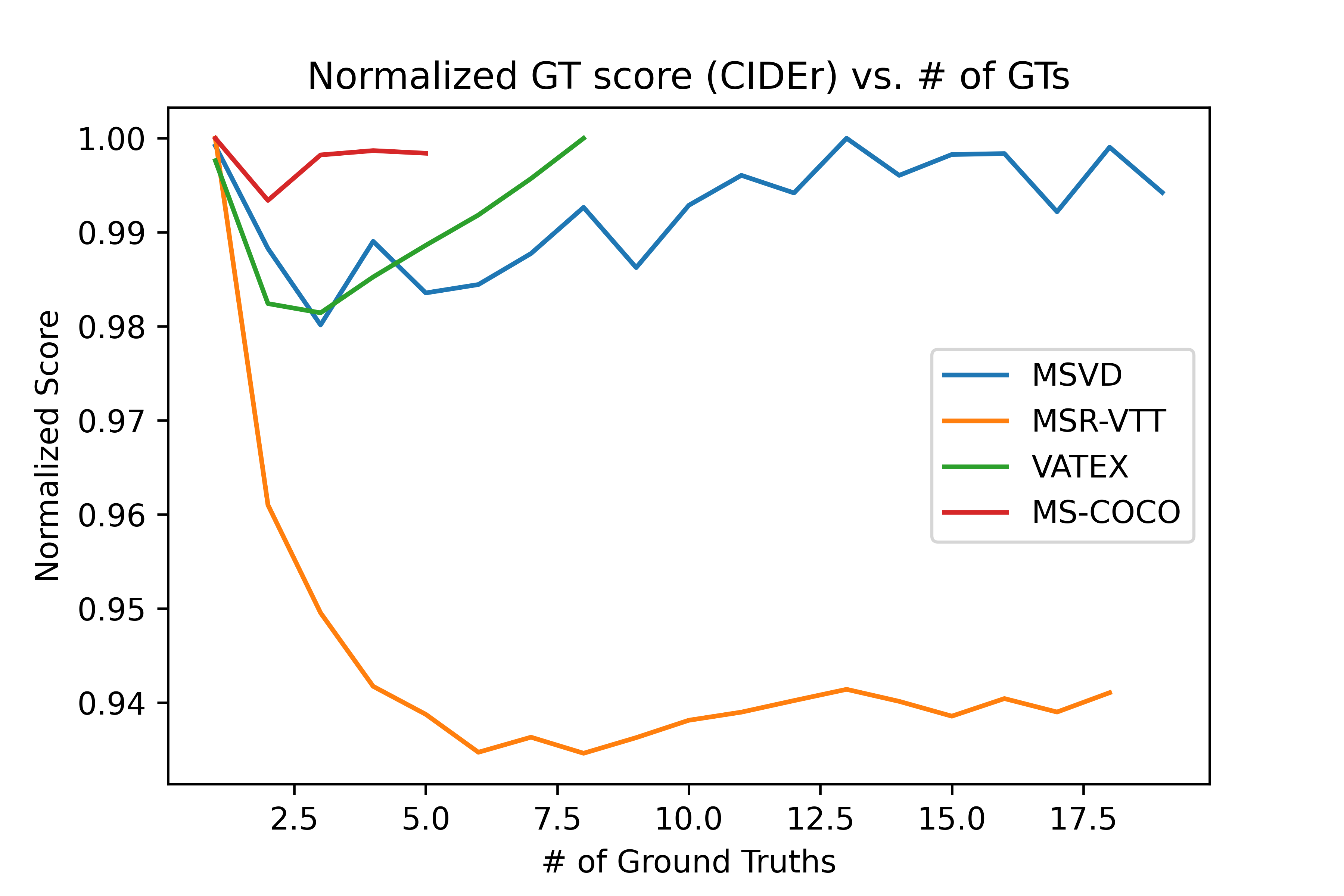}
    \label{fig:ngt_c}
\end{subfigure}
\caption{Performance of different metrics with respect to the number of ground truths considered in leave-one-out experiments. Raw scores are normalized to a maximum of 1, so we can compare the different datasets on the same plot.}
\label{fig:ngt}
\end{figure*}

To investigate how the number of ground truth metrics impacts the computation of the metrics, we performed several leave one out experiments as in \autoref{sec:loo} where we restricted the size of $R_i$ for each sample to a certain number of references $r$ by randomly sampling $r$ elements without replacement from the original reference set. This allows us to measure the approximate performance of the methods if the number of ground truths was reduced. The results of this experiment are given in \autoref{fig:ngt}. We can see from \autoref{fig:ngt} that except for CIDEr, increasing the number of ground truths increases the leave one out performance of the metrics. In fact, we can see that in most cases, the performance is nowhere near saturated, and collecting more ground truths will allow metrics to better capture the semantic variance of a scene. The standout among the group is CIDEr, in which the score does not increase as we increase the number of ground truths. This is primarily due to the IDF component of the CIDEr score, which penalizes increasing the number of tokens harshly. We can see that here, as we increase the number of ground truths, the CIDEr score \textit{decreases}! This suggests that CIDEr is relatively robust to adding more ground truth, however cannot capture as much semantic variance as the other metrics, as the CIDEr score does not materially account for new information from the ground truth samples.

\subsection{Concept-Diversity: Captions Required for BLEU Score}

One of the key experiments we perform is designed to measure the minimum number of captions from the training set that are required to ``solve" the test set of the dataset for a particular BLEU score. We first compute a set of all hypothesis descriptions from the training set. Then, for each sample in the test set, we compute the BLEU@4 score using that hypothesis for every sample in the test set. In the case of large datasets such as MS-COCO, which contains $591,435$ unique hypothesis captions, this can be time-consuming, even for the (relatively quick) BLEU@4 metrics. Each hypothesis thus has a score for each sample in the test dataset. Finding the minimal core-set of captions that covers this test dataset to a specified BLEU threshold is a weighted set-cover problem, which can be solved to an $O(\log N)$ approximation with a randomized rounding algorithm \citesupp{s-vazirani2001approximation}, however, we found that it was sufficient to use the greedy approximation algorithm for set cover, which selects the caption which covers the largest number of new samples at each iteration. Thus, the results in \autoref{fig:core_set} provide an upper bound on the possible number of captions required.

\autoref{fig:core_set} plots the required number of captions to achieve a BLEU@4 score of $X$ (the value on the X-axis) on every sample. Note that this requirement is \textit{more restrictive} than the plotted SOTA scores, which achieve a mean of $X$. Thus, the effect of this figure may be even more dramatic than is pictured. The reason for this discrepancy is we compute the core-set using a greedy set cover, and due to our implementation details, it is difficult to terminate the cover efficiently when a mean score is reached. 

While our work only computes the core-set for BLEU@4, we believe it would be interesting to see the numbers for other metrics, however, with current implementations, it may be intractable, as the computations require a full pairwise computation of the metrics between the hypotheses and the test-set samples. Additionally, metrics such as CIDEr which have dataset-wide effects would have to be estimated, requiring several hundred iterations of this experiment to achieve high-quality estimates of the performance. It thus remains interesting (and important) future work to explore how many captions are required to perform well on any given dataset for other metrics.

\subsection{Concept-Diversity: Feature Sets}

To measure the diversity of the datasets at a concept level, we look at how the ground truth captions overlap with the label sets from common feature extractors. If we find that this overlap is high, it suggests that features may have the ability to bias the model along the classification lines of the feature-extractor label set (since a lot of the time, the information extracted by the features is useful primarily for segmenting data along feature class boundaries).

\subsubsection{Computing Label-Set overlap}

We discuss two methods for computing label-set overlap in the main paper: exact match and fuzzy matching. Exact match is implemented as a string substring: i.e. does the label string appear as a direct substring of the caption. This method provides a lower bound on the true conceptual overlap, as it does not account for misspellings (which are surprisingly common in datasets such as MSR-VTT, and others collected using AMT without additional review steps), and other close matches. While this is a lower bound, it has the benefit of not introducing false-positive matches (as any match is guaranteed to be label overlap). We also discuss the use of fuzzy matching, which we implement using the fuzzywuzzy\footnote{\url{https://github.com/seatgeek/thefuzz}} library for approximate string matching with a threshold of 90. This library uses Levenshtein distance to compute approximate matching, however introduces false-positives which makes it difficult to analyze the overlap. In all cases, the numbers in \autoref{tab:overlap} represent the percentage of samples that have at least one reference description that has exact overlap with a label from the dataset.

We explore overlap on four common datasets for feature extraction: \\

    \paragraph{ImageNet-1K} \citesupp{s-deng2009imagenet} is a popular image classification dataset consisting of 1K labels for object classification ranging across a very wide variety of objects. We can see this from the overlap scores in \autoref{tab:overlap}, which are relatively high on almost all of the datasets. MSR-VTT is relatively low, suggesting that it is one of the most open-domain datasets among the datasets we explore. \\
    \paragraph{Kinetics-600} \citesupp{s-carreira2018short} is a popular dataset for action recognition, which contains 600 activities. We can see that the video datasets have a much higher overlap with kinetics, but even though MS-COCO is an image dataset only, there is still some overlap, suggesting that captions of static data still contain human inferences about motion and activity. \\
    \paragraph{MS-COCO} \citesupp{s-lin2014microsoft} is a dataset for object detection (and also for visual description) containing object-detection labels over 80 object classes from everyday life. Even though COCO has a relatively restricted object set, we can see that it consists of a set of very popular objects, as the overlap is more than 50\% for all captions. Additionally, it's interesting that the object labels for MS-COCO don't always appear in the captions themselves (as the self-overlap is only 92\%). \\
    \paragraph{Places-365} \citesupp{s-zhou2017places} is a dataset for scene recognition, consisting of 365 labels of scenes or settings for an image. We find empirically that the overlap for places is likely low, not due to a lack of descriptions of setting, but rather a lack of wide coverage of the variance of settings in Places. 
    
\subsubsection{Feature-Set Core-Sets and BLEU@4 performance}

To directly measure how transferable descriptions are along feature-extractor label axes, we explore the leave-one-out performance of captions sharing the same feature label, but from different samples in the dataset. The results of this experiment using BLEU@4 scores are given in \autoref{tab:features}. In order to compute the leave-one-out performance, we begin by computing a set of reference captions $R_c$ for each label in each feature-extractor label set, drawing from the training dataset. These concept-level reference sets consist of all captions containing that label as an exact sub-string. Then, for each sample $S_i$ with references $R_i$, we compute the set of all concepts overlapping that sample's references $C_i$. We then compute the hypothesis set for sample $S_i$ as $$H_i = \left[\bigcup_{c \in C_i} R_c\right] \big/ \{R_i\}$$ Next, for each hypothesis in $H_i$, we compute the BLEU@4 score for that hypothesis using ground truths $R_i$. The table \autoref{tab:features} reports the mean over all samples of the maximum across $H_i$ for each sample in the test set. The results of this metric are clear - when you use the best caption from another sample along feature boundaries, then these captions are relatively transferable (and almost always outperform samples from even the same sample).

\subsection{Tools \& Hardware}

The experiments in this paper are computed using the metric implementations provided by the MSCOCO evaluation toolkit in order to compute numeric metric values that are comparable with state of the art methods. In the experiments in the paper, we use the Stanford PTB\footnote{\url{https://nlp.stanford.edu/software/tokenizer.html}} tokenizer provided as part of the toolkit for tokenization and standardization. Unfortunately, because the MSCOCO toolkit does not explicitly specify a tokenization scheme and most works in video description do not subscribe to a standard tokenization tool, we are unable to be certain that the metric is consistent between our work, and the work presented in the state of the art papers.

The experiments are run in parallel on a machine with 96 AMD EPYC 7B12 cores and 378 GB of RAM running on Google Cloud Platform. Notably, the caption concept-overlap experiments require a very large amount of compute, with this machine requiring almost 10 hours to compute the BLEU score for the core-set concept overlap. We found scores such as METEOR \citesupp{s-agarwal2008meteor} and SPICE \citesupp{s-anderson2016spice} to be computationally prohibitive (requiring several months of sustained compute) for some of these experiments, thus, we do not include those scores in this work. We also do not report several modern metrics for this reason - as a major downside to many of the automated metrics that have recently been developed is their forward inference speed (up to 1000s of times slower than the computation of the BLEU score). A key area of future work is improving the computational performance of metrics, as this will allow such metrics to not only be used for more detailed analysis but will allow such metrics to be optimized directly using techniques such as self-critical sequence training \citesupp{s-rennie2017self}. 

\onecolumn
\clearpage
\pagebreak
\section{Additional Qualitative Examples}
\textit{\footnotesize{Additional qualitative examples are selected at random from the datasets using a random number generator over the length of each dataset. Some randomly selected samples are omitted due to explicit content in the visual data or descriptions (which is an additional cause for concern, but out of scope of the current research).}}

\begin{figure*}[!htbp]
    \centering
    \includegraphics[width=\linewidth]{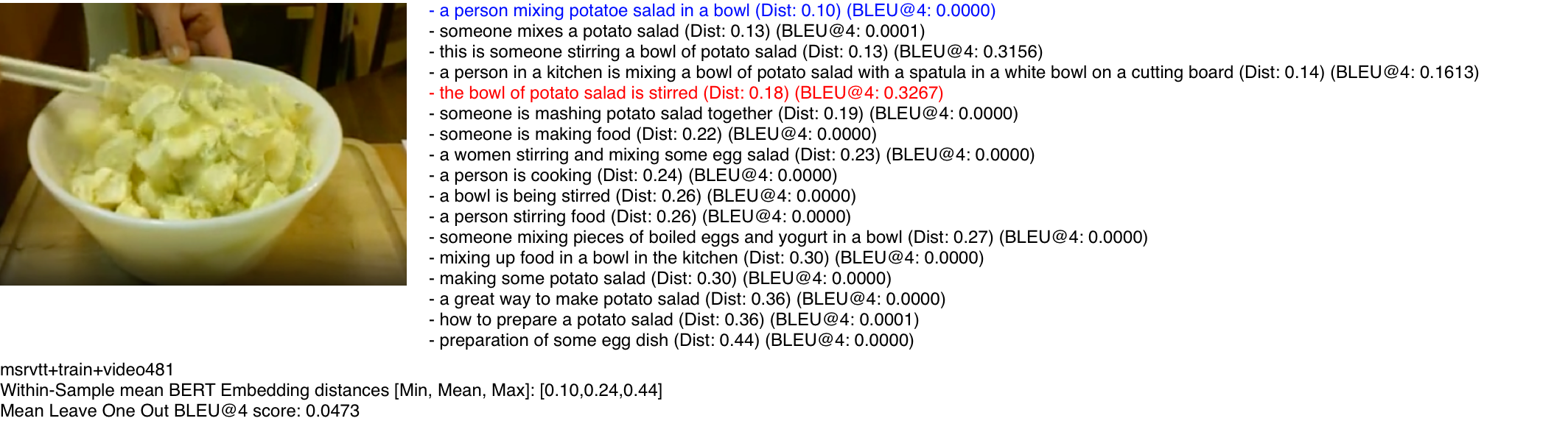}
    \caption{Qualitative example of metrics presented in the paper. The blue description is a description with the minimum distance from the sentence embedding mean, while the red description maximizes the mean BLEU@4 score to all other captions in the sample. Captions are ordered from top to bottom by similarity to the mean caption embedding (See section \ref{sec:h3}).}
    \label{fig:q2}
\end{figure*}

\begin{figure*}[!htbp]
    \centering
    \includegraphics[width=\linewidth]{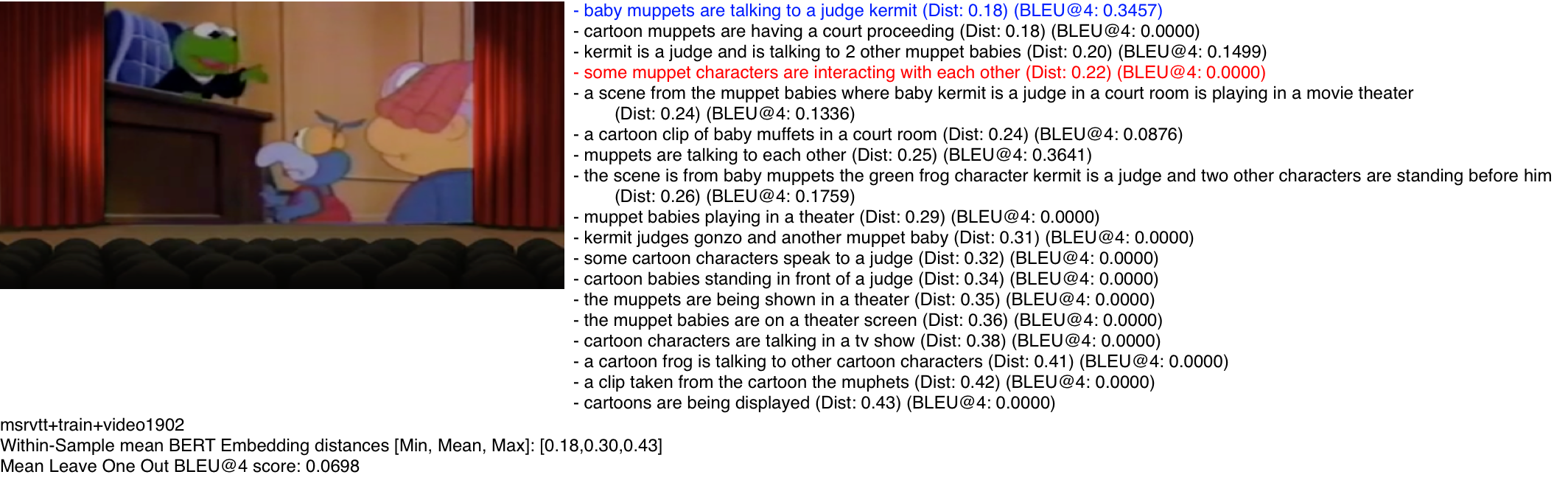}
    \caption{Qualitative example of metrics presented in the paper. The blue description is a description with the minimum distance from the sentence embedding mean, while the red description maximizes the mean BLEU@4 score to all other captions in the sample. Captions are ordered from top to bottom by similarity to the mean caption embedding (See section \ref{sec:h3}).}
    \label{fig:q3}
\end{figure*}

\begin{figure*}[!htbp]
    \centering
    \includegraphics[width=\linewidth]{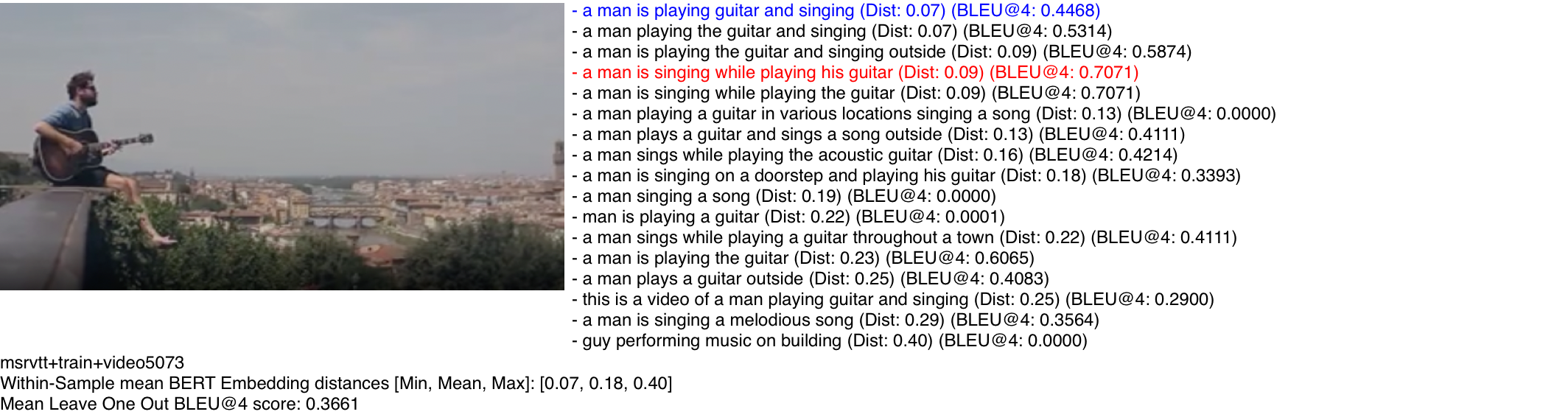}
    \caption{Qualitative example of metrics presented in the paper. The blue description is a description with the minimum distance from the sentence embedding mean, while the red description maximizes the mean BLEU@4 score to all other captions in the sample. Captions are ordered from top to bottom by similarity to the mean caption embedding (See section \ref{sec:h3}).}
    \label{fig:q4}
\end{figure*}


\begin{figure*}[!htbp]
    \centering
    \includegraphics[width=\linewidth]{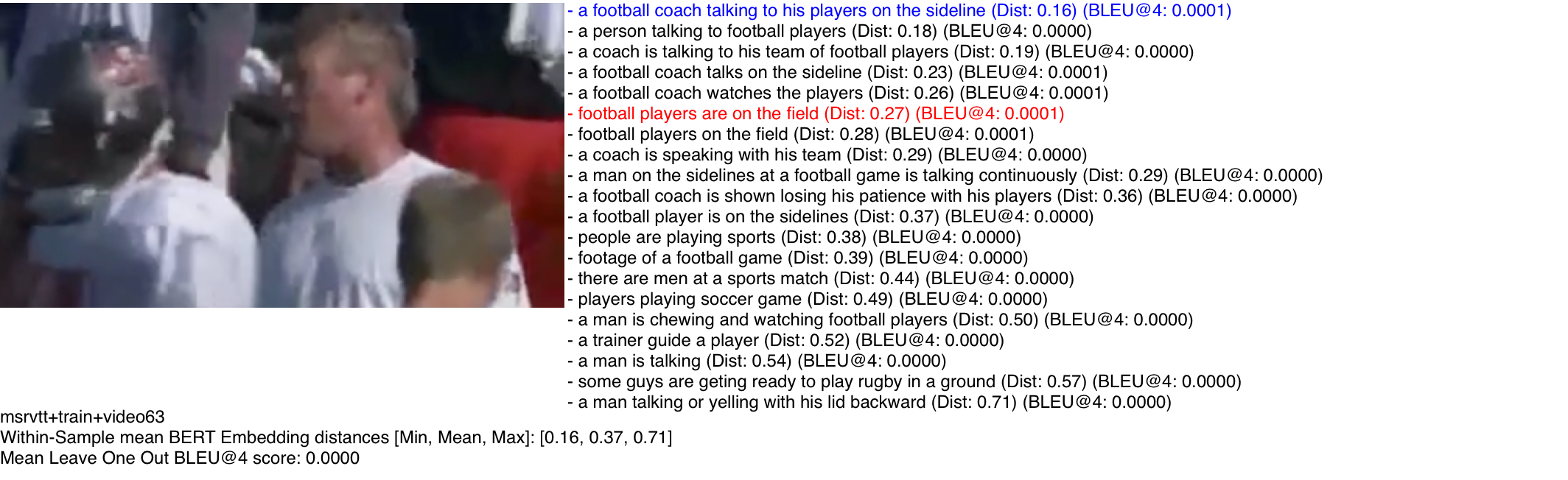}
    \caption{Qualitative example of metrics presented in the paper. The blue description is a description with the minimum distance from the sentence embedding mean, while the red description maximizes the mean BLEU@4 score to all other captions in the sample. Captions are ordered from top to bottom by similarity to the mean caption embedding (See section \ref{sec:h3}).}
    \label{fig:q6}
\end{figure*}

\begin{figure*}[!htbp]
    \centering
    \includegraphics[width=\linewidth]{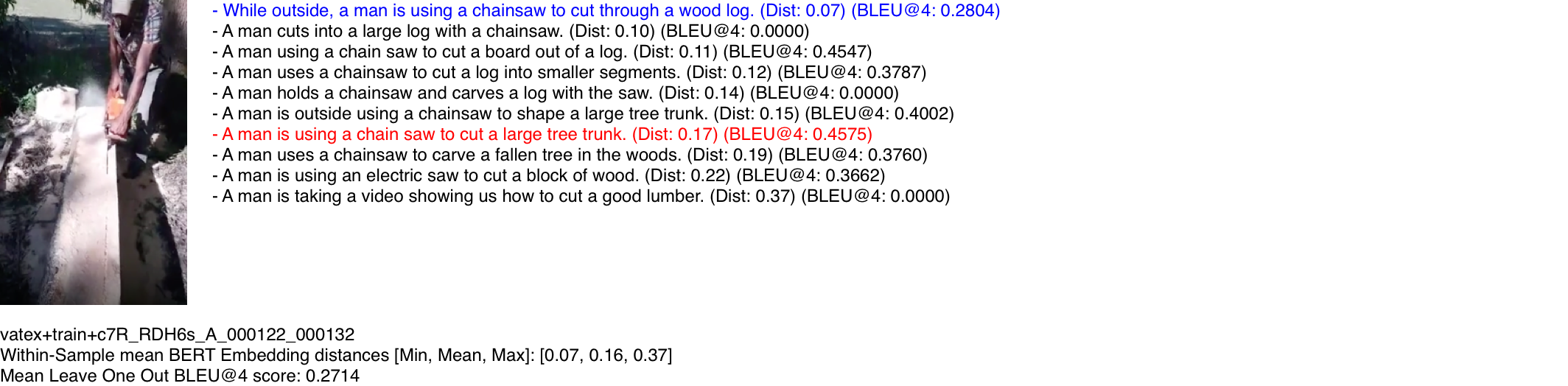}
    \caption{Qualitative example of metrics presented in the paper. The blue description is a description with the minimum distance from the sentence embedding mean, while the red description maximizes the mean BLEU@4 score to all other captions in the sample. Captions are ordered from top to bottom by similarity to the mean caption embedding (See section \ref{sec:h3}).}
    \label{fig:q7}
\end{figure*}

\begin{figure*}[!htbp]
    \centering
    \includegraphics[width=\linewidth]{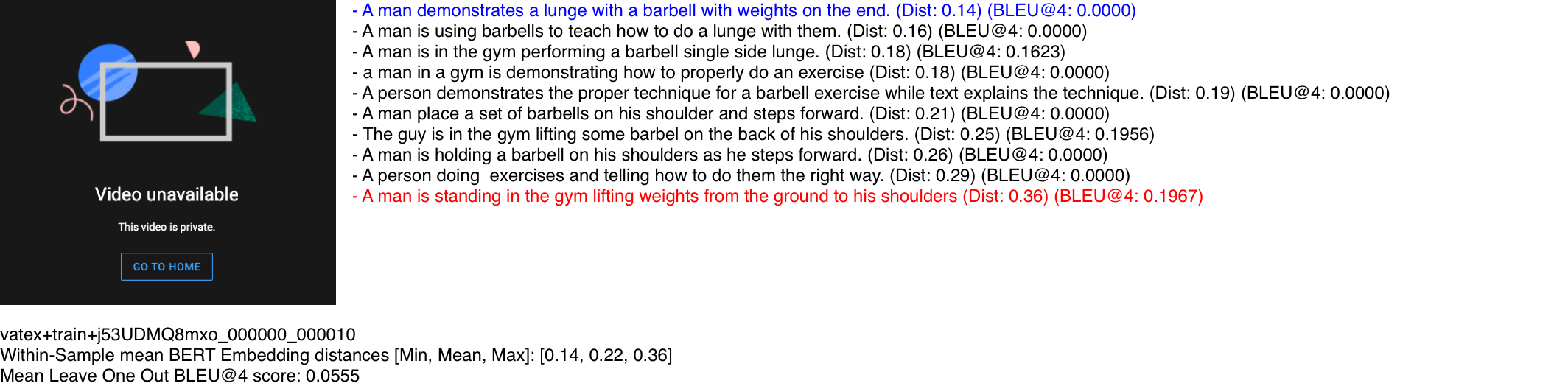}
    \caption{Qualitative example of metrics presented in the paper. The blue description is a description with the minimum distance from the sentence embedding mean, while the red description maximizes the mean BLEU@4 score to all other captions in the sample. Captions are ordered from top to bottom by similarity to the mean caption embedding (See section \ref{sec:h3}). The visual content of this video is missing (as the video has become private since the collection of the dataset), however we include the video as it is one of the randomly sampled instances.}
    \label{fig:q8}
\end{figure*}

\begin{figure*}[!htbp]
    \centering
    \includegraphics[width=\linewidth]{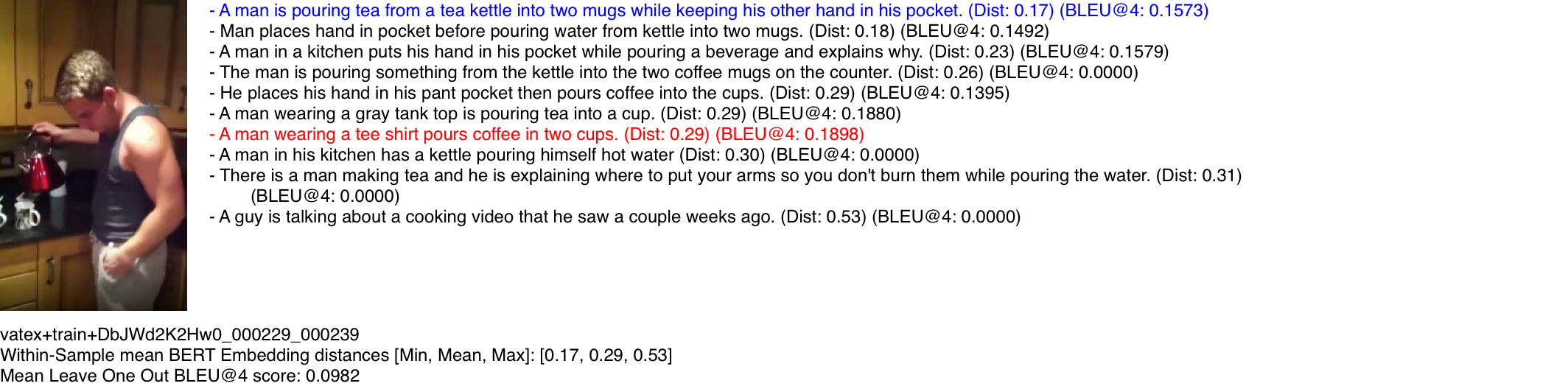}
    \caption{Qualitative example of metrics presented in the paper. The blue description is a description with the minimum distance from the sentence embedding mean, while the red description maximizes the mean BLEU@4 score to all other captions in the sample. Captions are ordered from top to bottom by similarity to the mean caption embedding (See section \ref{sec:h3}).}
    \label{fig:q9}
\end{figure*}

\begin{figure*}[!htbp]
    \centering
    \includegraphics[width=\linewidth]{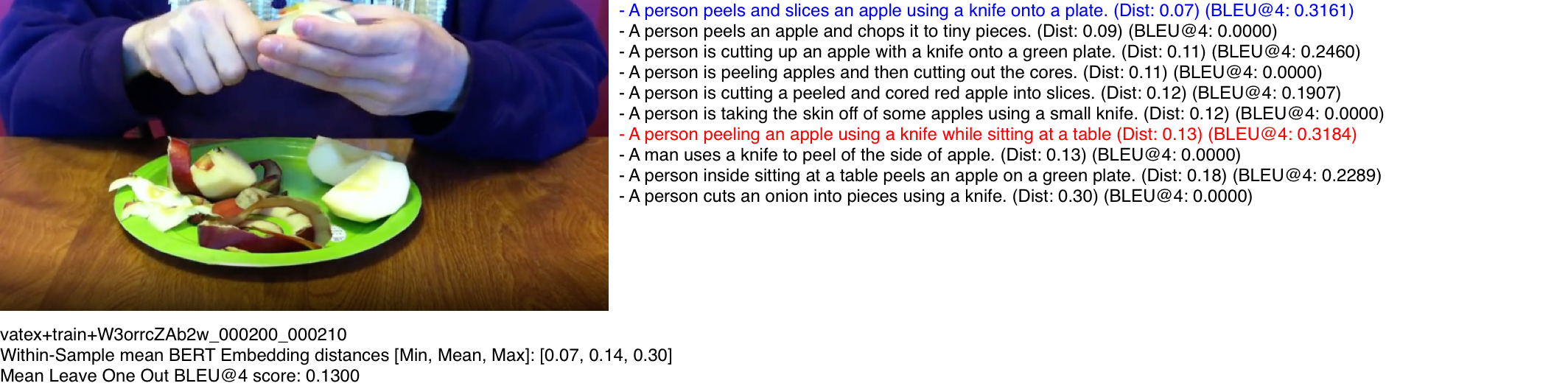}
    \caption{Qualitative example of metrics presented in the paper. The blue description is a description with the minimum distance from the sentence embedding mean, while the red description maximizes the mean BLEU@4 score to all other captions in the sample. Captions are ordered from top to bottom by similarity to the mean caption embedding (See section \ref{sec:h3}).}
    \label{fig:q10}
\end{figure*}

\begin{figure*}[!htbp]
    \centering
    \includegraphics[width=\linewidth]{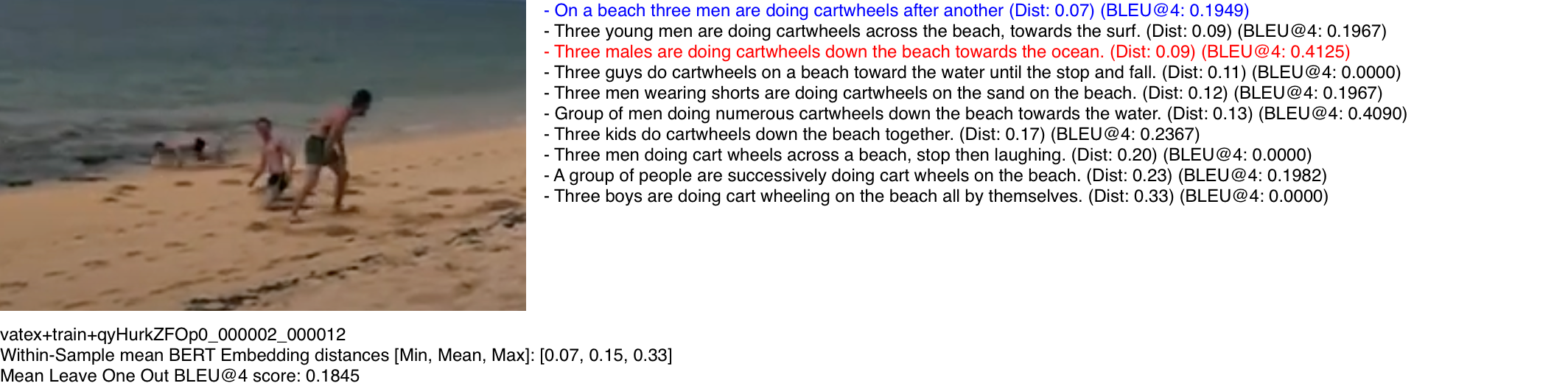}
    \caption{Qualitative example of metrics presented in the paper. The blue description is a description with the minimum distance from the sentence embedding mean, while the red description maximizes the mean BLEU@4 score to all other captions in the sample. Captions are ordered from top to bottom by similarity to the mean caption embedding (See section \ref{sec:h3}).}
    \label{fig:q11}
\end{figure*}

\begin{figure*}[!htbp]
    \centering
    \includegraphics[width=\linewidth]{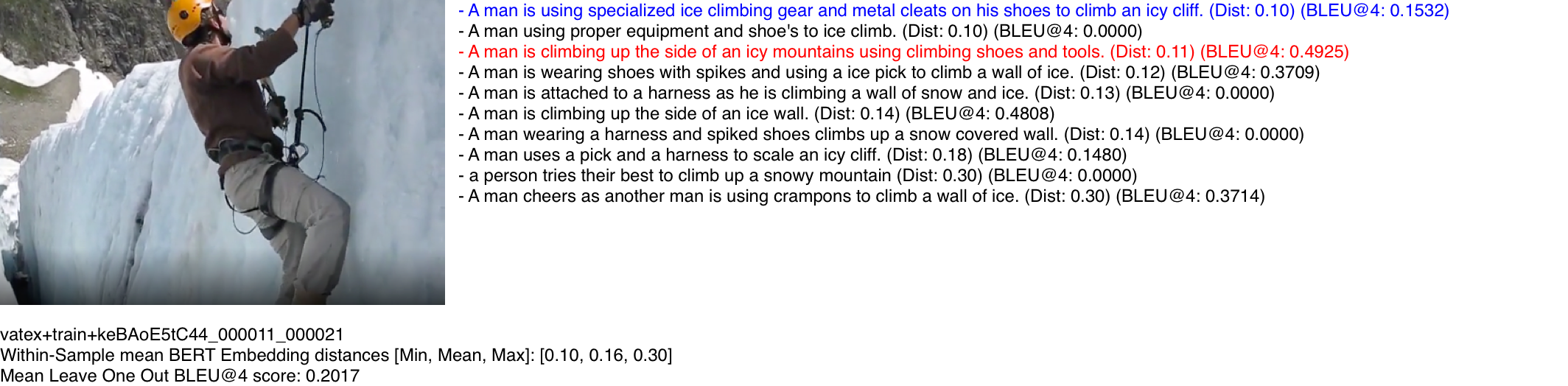}
    \caption{Qualitative example of metrics presented in the paper. The blue description is a description with the minimum distance from the sentence embedding mean, while the red description maximizes the mean BLEU@4 score to all other captions in the sample. Captions are ordered from top to bottom by similarity to the mean caption embedding (See section \ref{sec:h3}).}
    \label{fig:q12}
\end{figure*}

\begin{figure*}[!htbp]
    \centering
    \includegraphics[width=\linewidth]{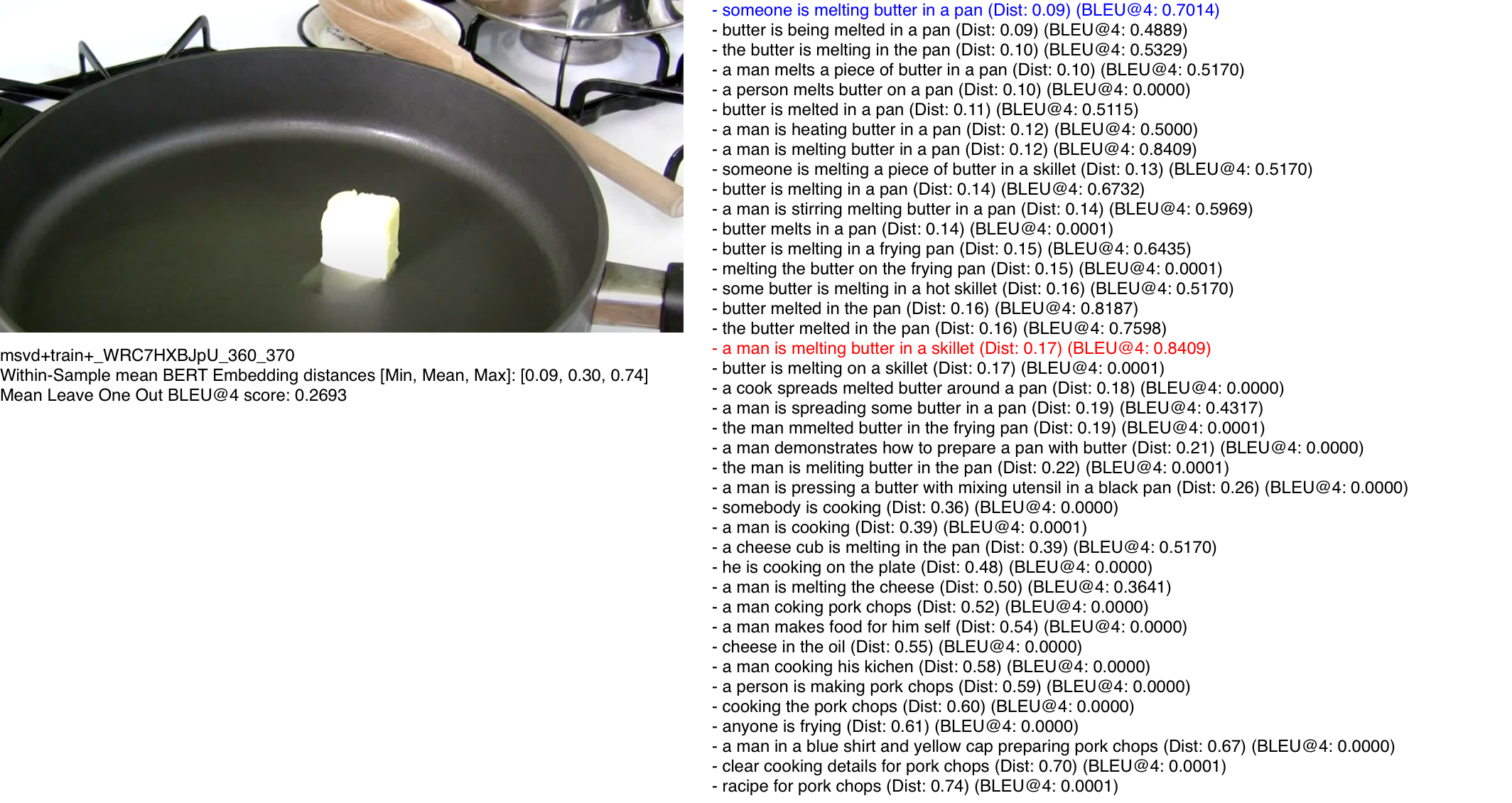}
    \caption{Qualitative example of metrics presented in the paper. The blue description is a description with the minimum distance from the sentence embedding mean, while the red description maximizes the mean BLEU@4 score to all other captions in the sample. Captions are ordered from top to bottom by similarity to the mean caption embedding (See section \ref{sec:h3}).}
    \label{fig:q13}
\end{figure*}

\begin{figure*}[!htbp]
    \centering
    \includegraphics[width=\linewidth]{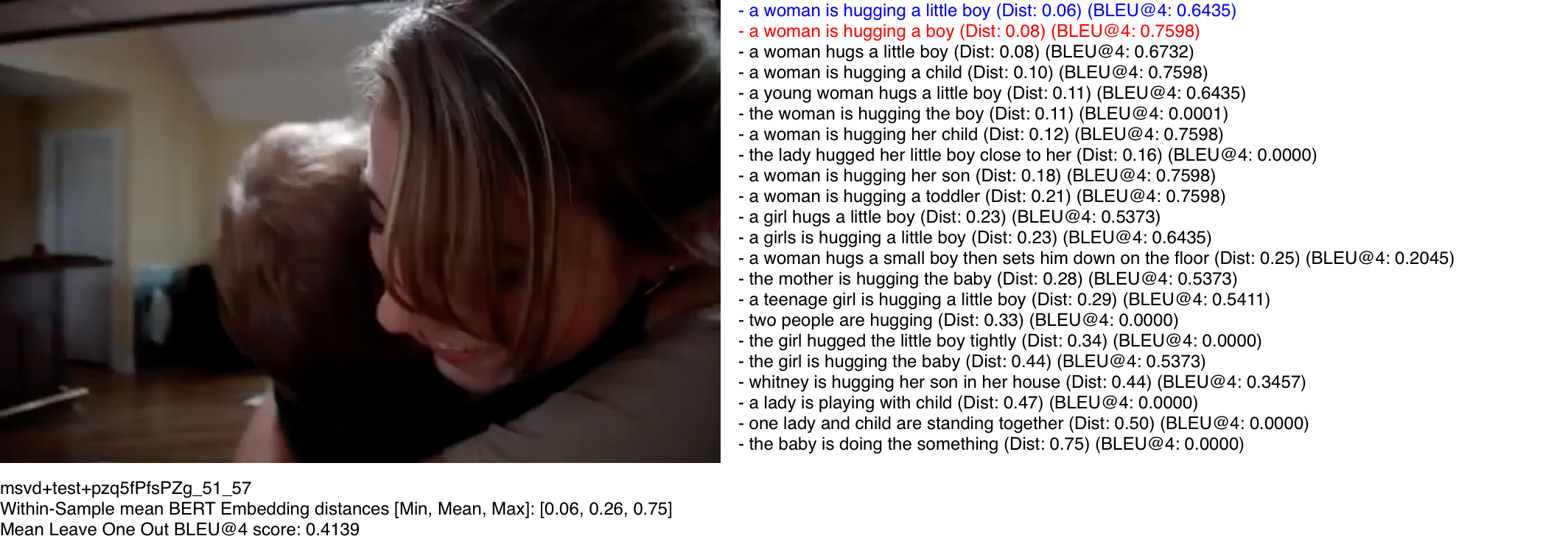}
    \caption{Qualitative example of metrics presented in the paper. The blue description is a description with the minimum distance from the sentence embedding mean, while the red description maximizes the mean BLEU@4 score to all other captions in the sample. Captions are ordered from top to bottom by similarity to the mean caption embedding (See section \ref{sec:h3}).}
    \label{fig:q14}
\end{figure*}

\begin{figure*}[!htbp]
    \centering
    \includegraphics[width=\linewidth]{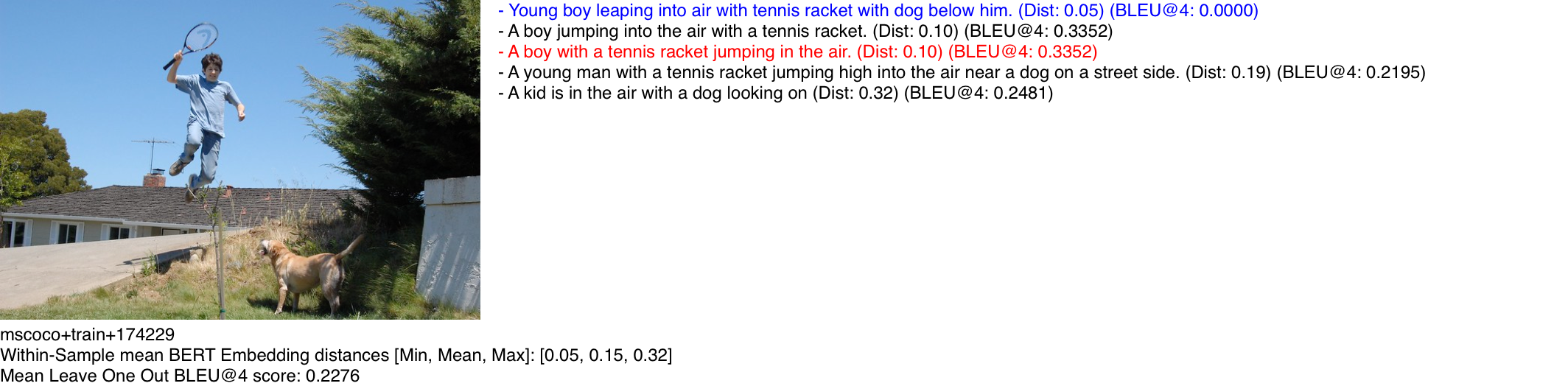}
    \caption{Qualitative example of metrics presented in the paper. The blue description is a description with the minimum distance from the sentence embedding mean, while the red description maximizes the mean BLEU@4 score to all other captions in the sample. Captions are ordered from top to bottom by similarity to the mean caption embedding (See section \ref{sec:h3}).}
    \label{fig:q15}
\end{figure*}

\begin{figure*}[!htbp]
    \centering
    \includegraphics[width=\linewidth]{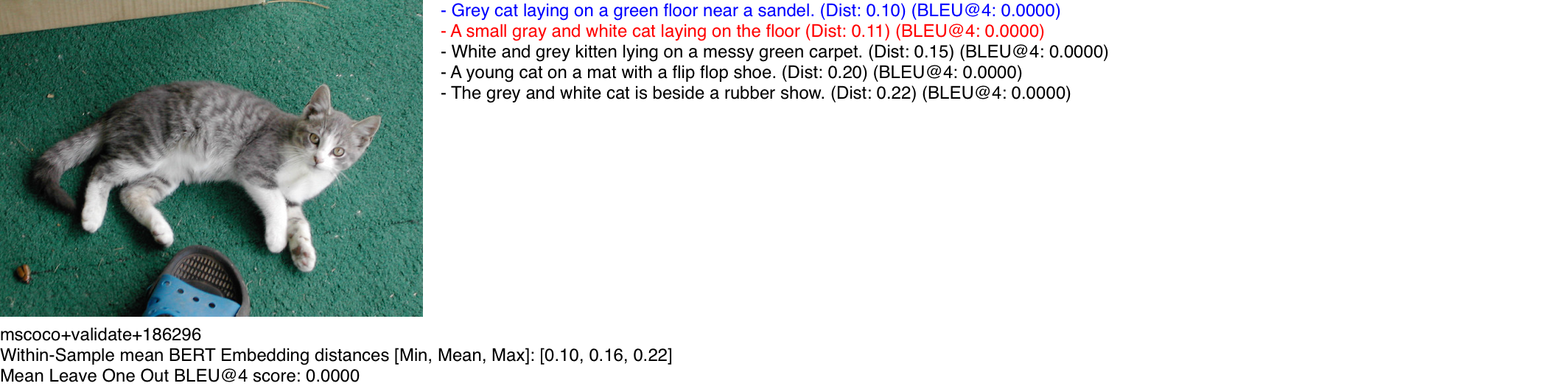}
    \caption{Qualitative example of metrics presented in the paper. The blue description is a description with the minimum distance from the sentence embedding mean, while the red description maximizes the mean BLEU@4 score to all other captions in the sample. Captions are ordered from top to bottom by similarity to the mean caption embedding (See section \ref{sec:h3}).}
    \label{fig:q16}
\end{figure*}